\documentclass[lettersize,10pt,journal]{IEEEtran}

\usepackage{algorithmic}
\usepackage{algorithm}
\usepackage{multirow}%
\usepackage{amsmath,amssymb,amsfonts}%
\usepackage{subfig}
\usepackage{xcolor}
\usepackage{pdfpages}
\usepackage{dirtytalk}
\usepackage{listings}%
\usepackage[nolist,nohyperlinks]{acronym}
 \begin{acronym}
    \acro{TS}{Thompson sampling}
    \acro{MAB}{multi-armed bandit}
    \acro{SRTS}{Sharpe ratio Thompson sampling}
    \acro{MM}{misallocation minimization}
    \acro{BAI}{best arm identification}
    \acro{GBM}{geometric Brownian motion}
    \acro{MDP}{Markov decision process}
    \acro{RL}{reinforcement learning}
    \acro{MPT}{modern portfolio theory}
    \acro{OTC}{over-the-counter}
    \acro{UCB}{upper confidence bound}
    \acro{TS}{Thompson sampling}
    \acro{i.i.d}{independent and identically distributed}
    \acro{CVaR}{conditional value-at-risk}
    \acro{VaR}{value-at-risk}
    \acro{SR}{Sharpe ratio}
 \end{acronym}

\usepackage{subfig}
\usepackage{xcolor}
\usepackage{float}
\usepackage{svg}
\usepackage{pdfpages}
\usepackage{comment}
\usepackage{graphicx,wrapfig,lipsum}
\usepackage{rotating}
\usepackage{amsmath, amssymb, amsthm}
\usepackage{stmaryrd}
\usepackage{tikz}
\usepackage{verbatim}
\usepackage{url}
\usepackage[utf8]{inputenc}
\usepackage[english]{babel}

\newtheorem{theorem}{Theorem}[]
\newtheorem{corollary}{Corollary}[]

\newtheorem{lemma}[]{Lemma}

\newtheorem{assumption}{Assumption}[]

\usepackage{multicol}
\usepackage{scalerel}
\usepackage{multicol}
\usepackage{setspace}
\newtheorem{definition}{Definition}
\usepackage[noadjust]{cite}
\usepackage{booktabs}
\usepackage{bbm}
\usepackage[normalem]{ulem}
\usepackage{color}
\usepackage{dsfont}
\usepackage{bm}
\usepackage{setspace}
\usepackage{mathtools}
\usepackage{csquotes}
\usetikzlibrary{automata}
\usetikzlibrary{arrows}
\usetikzlibrary{shapes.geometric, arrows}
\usepackage[]{footmisc}
\usetikzlibrary{positioning}
\usetikzlibrary{calc}
\usetikzlibrary{arrows}
\allowdisplaybreaks

\usepackage[margin=0.5cm]{caption}
\makeatletter
\newcommand{\vast}{\bBigg@{4}}
\newcommand{\Vast}{\bBigg@{5}}
\makeatother

\usepackage{accents}

\DeclareMathOperator*{\argmax}{argmax}

\DeclarePairedDelimiterX{\infdivx}[2]{(}{)}{%
  #1\;\delimsize\|\;#2%
}

\begin{document}

\title{Variance-Optimal Arm Selection: Misallocation Minimization and Best Arm Identification}

\author{
\IEEEauthorblockN{Sabrina~Khurshid}, {\it Graduate Student Member, IEEE}, \IEEEauthorblockN{Gourab~Ghatak}, {\it Senior Member, IEEE}, Mohammad~Shahid~Abdulla
\thanks{Sabrina Khurshid and G. Ghatak are with the Department of Electrical Engineering, IIT Delhi, New Delhi, India, 110016; Email: eez218683@ee.iitd.ac.in,  gghatak@ee.iitd.ac.in. M.S.Abdulla is with the Information Systems, IIM Kozhikiode, Kerela, India, 110020; Email: shahid@iimk.ac.in}
\vspace{-0.5cm}}

\maketitle
\begin{abstract}
This paper focuses on selecting the arm with the highest variance from a set of $K$ independent arms. Specifically, we focus on two settings: (i) misallocation minimization setting, that penalizes the number of pulls of suboptimal arms in terms of variance, and (ii) fixed-budget best arm identification setting, that evaluates the ability of an algorithm to determine the arm with the highest variance after a fixed number of pulls. We develop a novel online algorithm called \texttt{UCB-VV} for the \ac{MM} and show that its upper bound on misallocation for bounded rewards evolves as $\mathcal{O}\left(\log{n}\right)$ where $n$ is the horizon. By deriving the lower bound on the misallocation, we show that \texttt{UCB-VV} is order optimal. For the fixed budget \ac{BAI} setting we propose the \texttt{SHVV} algorithm. We show that the upper bound of the error probability of \texttt{SHVV} evolves as $\exp\left(-\frac{n}{\log(K) H}\right)$,  where $H$ represents the complexity of the problem, and this rate matches the corresponding lower bound. We extend the framework from bounded distributions to sub-Gaussian distributions using a novel concentration inequality on the sample variance and standard deviation. Leveraging the same, we derive a concentration inequality for the empirical Sharpe ratio (SR) for sub-Gaussian distributions, which was previously unknown in the literature. Empirical simulations show that \texttt{UCB-VV} consistently outperforms \texttt{$\epsilon$-greedy} across different sub-optimality gaps though it is surpassed by \texttt{VTS}, which exhibits the lowest misallocation, albeit lacking in theoretical guarantees. We also illustrate the superior performance of \texttt{SHVV}, for a fixed budget setting under 6 different setups against uniform sampling. Finally, we conduct a case study to empirically evaluate the performance of the \texttt{UCB-VV} and \texttt{SHVV} in call option trading on $100$ stocks generated using \ac{GBM}.
\end{abstract}


\section{Introduction and Motivation}
Historically, variance has served as the first and foundational risk measure in portfolio optimization. The modern portfolio theory is based upon the classical Markowitz model, which also employs variance as a measure of risk \cite{{kalymon1971estimation}, {hult2012risk},{kim2012measuring}}. In economic models, the decision-makers account for uncertainty and variance to maintain robustness \cite{hansen2001robust}. In decision-making and optimization, identifying and mitigating sources of high variance is crucial to ensuring stability and robustness in outcomes \cite{{suttonreinforcement}, {avramidis1996integrated}, {koivu2005variance}}.

Recent advances in risk-sensitive decision making involve more sophisticated risk measures such as \ac{VaR} and \ac{CVaR} \cite{{chang2022unifying}, {cassel2018general}, {tamkin2019distributionally}, {tan2023cvar}, {liang2023distribution}}. Unlike \ac{VaR} and \ac{CVaR}, which rely on tail-specific assumptions and involve complex computations related to probability thresholds and conditional expectations, variance is distribution-agnostic and computationally lightweight. From the perspective of selecting the arm with the highest variance, we study two settings of the \ac{MAB} framework - \ac{MM} and \ac{BAI}. Specifically, the \ac{MM} setting attempts to minimize the number of pulls of sub-optimal arms in terms of the variance. On the contrary, the \ac{BAI} setting attempts to determine the arm with the highest variance with a high probability. The fixed-budget \ac{BAI} focuses on identifying the best arm with minimum error for a fixed number of samples~\cite{{audibert2010best}, {carpentier2016tight},{karnin2013almost}}. This is particularly relevant in scenarios like A/B testing, clinical trials, or recommendation systems, where the ultimate goal is to find the single best option.

Variance has been central to research on asset pricing and portfolio management. High-variance assets introduce instability and expose portfolios to erratic swings, requiring their identification and possible exclusion. For portfolio managers, maximizing the frequency of selecting the highest variance option unlocks opportunities to optimize returns while leveraging the asymmetrical payoff structure of call options from favorable price movements. For example, for a call option buyer, a higher variance is particularly advantageous as it increases the potential for substantial gains while capping the maximum loss to the premium paid for the option. In this paper, we develop bandit algorithms  for identification of arm with the highest variance in \ac{MM} and \ac{BAI} settings. It must be noted that the results of our framework can be extended to the setting to pick the arm with the minimum variance as well.

\subsection{Related work}
\subsubsection{Literature in bandit setting}
In the regret setting, \cite{audibert2009exploration} investigates \texttt{UCB-V}, a variant of the \ac{UCB} algorithm for stochastic \ac{MAB} that incorporates empirical variance estimates of arm payoffs to pick the arm with the highest mean. The algorithm performs better than traditional methods, especially when suboptimal arms have low variance. However, it depends on an unavoidable upper-limit $u$, which can reduce efficiency when $u$ is large but actual variances are small. The authors in \cite{mukherjee2018efficient} present \texttt{E-UCBV}, a variance-aware \ac{UCB} algorithm that achieves tighter regret bounds than \texttt{UCB1}. By incorporating efficient arm elimination, \texttt{E-UCBV} demonstrates superior performance in scenarios with small gaps and low-variance suboptimal arms. However, the algorithm has notable limitations, including high complexity in regret bounds and constraints on $n$. The authors in \cite{liu2020risk, zhu2020thompson} address risk-aware decision-making and design algorithms for mean-variance \ac{MAB}. 

The \ac{BAI} problem has been extensively investigated across a broad range of settings, e.g, for linear bandits \cite{jedra2020optimal}, cost-aware \ac{BAI} \cite{qin2025cost}, as well as risk-averse bandits \cite{kagrecha2022statistically}. In risk-averse \ac{BAI} problem, \cite{david2018pac} considers \ac{VaR}-optimization. Also, \cite{kagrecha2022statistically}, \cite{kagrecha2019distribution},  optimizes a mean-CVaR framework within a fixed budget framework. They propose distribution oblivious algorithms, i.e, they do not require prior knowledge of the arms' underlying distributions and are asymptotically near-optimal. However, these algorithms come with weaker performance guarantees compared to those that leverage prior knowledge of the arm distributions, highlighting an inherent trade-off between robustness and efficiency. Variance optimization, in particular, often necessitates stronger assumptions about the reward distributions (e.g., sub-Gaussianity) to ensure accurate and reliable estimation. Without such assumptions, variance estimates can become highly sensitive to noise and hence unreliable.

\subsubsection{Literature in finance}
In the context of finance, variance is used to model risk premium, providing insights into market sentiment and investor behavior. Specifically, the variance risk premium (a compensation that investors demand for bearing the risk associated with the volatility in the returns of an asset) plays a critical role in both theoretical and practical applications. This concept has been leveraged to develop trading strategies, such as selling volatility when the premium is high, as demonstrated in \cite{{carr2009variance},{zhou2018variance}}. Beyond its application in trading strategies, variance is also instrumental in understanding cross-market dynamics. For instance, \cite{hamao1990correlations} investigated how volatility in one market, such as Japan, influences volatility in another, such as the U.S., using variance as a metric to assess volatility spillovers. Similarly, \cite{engle1988meteor} examined intraday volatility in foreign exchange markets, employing variance to model the clustering of volatility over time. This clustering, a manifestation of the heteroscedastic behavior, is effectively captured by variance. \cite{french1987expected} explored the relationship between expected stock returns and volatility, with volatility quantified by variance, thereby informing optimal portfolio construction and asset pricing.  Variance also plays a critical role in strategies involving trading call options, as higher variance increases the likelihood of the underlying asset’s price exceeding the strike price, thereby enhancing the intrinsic value of the option {\cite{hull2021options}. Also, it is a well-known fact that a call option on a volatile stock is more valuable than one that is less volatile \cite{merton1978returns}. Collectively, these studies illustrate the multifaceted role of variance in financial research and practice. 

\subsection{Contributions and Organization}
     \begin{enumerate}
    \item {\bf Misallocation algorithm and optimality:} First, for the \ac{MM} setting, we propose an algorithm -- \texttt{UCB-VV} that maximizes the frequency of selecting the arm with the highest variance. We derive an upper bound of order $\mathcal{O}(\log n)$ on the variance-misallocation of \texttt{UCB-VV}, together with a matching information-theoretic lower bound, thereby establishing the order optimality of the proposed algorithm for variance-based \ac{MM}. These results are presented in Section~II.
    \item {\bf \ac{BAI} algorithm and optimality:} Next, we propose an arm-elimination based algorithm called \texttt{SHVV} to identify the arm with the highest variance within a fixed budget of pulls.  Similar to the misallocation case, we derive matching upper and lower bounds on the probability of error for highest-variance identification and prove that \texttt{SHVV} is order optimal. These results are presented in Section~III. \\
    {\bf Key novelty in 1) and 2):} This is the first work that develops algorithms for selecting the arm with the highest variance and proves their order optimality. Derivation of all core probabilistic components for the variance objective is required even for \ac{BAI} algorithms. Consequently, we resort to tools like McDiarmid's inequality, Bernstein-type bound and KL-divergence for variance to establish sharp deviation guarantees for the sample variance.
    \item {\bf Extension to sub-Gaussian distributions and applications to \ac{SR}:} While the main theoretical results are derived for bounded rewards, we extend the framework to sub-Gaussian distributions. For this, we derive a novel Bernstein-type concentration inequality for the sample variance and standard deviation of a sub-Gaussian distribution. This extension is technically nontrivial because the square of a sub-Gaussian variable is sub-exponential, necessitating a two-term tail bound that cannot be obtained by direct extension of mean-based analyses. Leveraging this result, we derive a new, path-independent concentration inequality for the empirical \ac{SR} statistic. We further demonstrate how this bound enables the construction of a \ac{UCB}-type algorithm for \ac{SR} maximization and provide a sketch of its regret analysis. These results are presented in Section~IV. \\
    {\bf Key novelty in 3):} To the best of our knowledge, this is the first work that provides a path-independent concentration bound (unlike \cite{khurshid2025optimizing}) for the sample \ac{SR} for sub-Gaussian distributions. The resulting algorithm of \ac{SR} optimality and its regret bound has not been reported in the literature.
    \item {\bf Empirical evaluation and financial case study:} 
    Finally, following the theoretical development, in Section~V, we empirically evaluate the proposed algorithms by comparing \texttt{UCB-VV} with baseline algorithms such as \texttt{$\epsilon$-greedy} and \texttt{VTS}. We also demonstrate the superior performance of \texttt{SHVV} for a fixed budget setting under 6 different setups. Then, in Section~VI,
    we validate the efficacy of the proposed algorithms (both \texttt{UCB-VV} and \texttt{SHVV}) on a real-world application -- we present a call option trading case study based on \ac{GBM}-simulated stock prices as a qualitative financial illustration of variance driven decision making under convex payoffs.
\end{enumerate} 
 
\section{Problem formulation for misallocation}
\label{sec:2}
Consider a \ac{MAB} framework with $K\geq 2$ arms, a single player, and a time horizon $n$. We assume that the underlying rewards distributions of the arms are $\nu_1,\nu_2,\dots,\nu_K$, each of which is bounded\footnote{In Section~\ref{sec:4}, we relax the boundedness assumption and demonstrate the extension of our framework to sub-Gaussian distributions.} in $[l,u]$. The means, $\{\mu_i\}$, and the variances, $\{\sigma_i^2\}$, of the distributions are fixed but unknown to the player. When playing an arm $i$ at time $t$, a random reward $X_i(t) \sim \nu_i$ is received. The reward samples are independent across arms and time slots. A policy $\pi$ orchestrates the action of the player at each time instant and is driven by the historical data. Let $s_i(n)$ represent the number of plays of an arm $i$ until (and including) time $n$ by the player. 
Without loss of generality, we assume that the optimal arm is unique and $\sigma^2_{\ast} = \max_{i \in [K]} \sigma_i^2$.

\subsection{Preliminaries for upper bound}
\begin{definition}
\label{Definition:1}
The unbiased empirical variance of an arm $i$ at time $t$, calculated from the i.i.d. samples $\{X_i(s
_i(q))\}_{q=1}^t$ drawn from its distribution $\nu_i$ is given as
\begin{align}
 \bar{V}_i(t) = \frac{1}{s_{i}(t)-1} \sum_{q=1}^{s_{i}(t)} \big(X_i(s_i(q)) - \bar{\mu}_i(s_i(t))\big)^2,
    \label{eq:sample_v}
\end{align}
where  $ \bar{\mu}_i(s_i(t)) = \frac{1}{s_i(t)} \sum_{q=1}^{s_i(t)} X_i(s_i(q)).$
\end{definition}

\begin{definition}
Let $\sigma^2_{\ast}$ be the variance-optimal arm. The instantaneous loss at time $t$ is $ \ell_t = \mathbb{I}\{ I_t \neq * \}$. The cumulative misallocation up to horizon $n$ is
\begin{align}
    \mathcal{M}(n) = \sum_{i \ne *} s_i(n) = \mathbb{E}\left[\sum_{t=1}^{n} \mathbb{I}\{ I_t \neq * \}\right]
\end{align}
\label{eq:miss_definition}
The misallocation rate is 
\begin{align*}
\mathfrak{m}(n)
= \frac{1}{n} \mathcal{M}(n) =\frac{1}{n}\,\mathbb{E}\left[\sum_{t=1}^{n} \mathbb{I}\{ I_t \neq * \}\right]
\end{align*}
\end{definition}
Unlike mean-reward maximization where the loss at time $t$ depends on the sub-optimality gap, variance maximization does not associate a meaningful incremental penalty with the magnitude of the variance gap. Once an arm is known to have lower variance than the optimal arm, any pull of that arm is equally undesirable, regardless of how small or large the gap may be. Consequently,  the performance criterion naturally reduces to a $[0,1]$-type loss: at each time step, the learner either allocates the pull to the variance-optimal arm (no loss) or to a suboptimal arm (unit loss). This leads directly to the misallocation measure introduced in Definition~\ref{eq:miss_definition}, which evaluates how frequently the learner selects a suboptimal arm. This metric differs fundamentally from both cumulative regret which measures how bad each mistake is, and fixed-budget BAI which evaluates only the terminal recommendation. The misallocation rate instead captures the learner's anytime behavior and penalizes every suboptimal selection during the learning process. It therefore provides an interpretable, gap-independent measure suited to the identification driven nature of variance maximization.

All algorithms and theoretical guarantees in this paper are based on the unbiased sample variance estimator defined in (\ref{eq:sample_v}). Note that employing an biased estimate of variance does not have any impact on our guarantees. As $t$ increases, the distinction between $t-1$ and $t$ becomes increasingly negligible due to their asymptotic equivalence. This adjustment impacts all arms equally without altering their relative ordering.

\subsection{\texttt{UCB-VV} Algorithm}
We introduce \texttt{UCB-VV} (presented in Algorithm \ref{alg:algorithm_vv}) to pull an arm with the highest variance most frequently. The algorithm pulls the arm with the highest index $B_{i}(t)_{\rm VV}$, where
\begin{align}
    B_{i}(t)_{\texttt{VV}} = \bar{V}_i(t) + \sqrt{\frac{2\log{t}} {s_i}},
    \label{eq:algo_VV}
\end{align}
comprises of two distinct components: $\bar{V}_i(t)$, represents the estimated variance, while the second term, $C_{t,s_i} \triangleq {\sqrt{\frac{2\log{t}} {s_i(t)}}}$, corresponds to its associated confidence bound.

\subsection{Upper bound on misallocation of \texttt{UCB-VV}} \label{sec3}
First, we employ McDiarmid’s inequality to derive a concentration bound for the variance estimator $\bar{V}(n)$. Based on this, we derive an upper bound on misallocation of \texttt{UCB-VV} and show it evolves as $\mathcal{O}\log(n)$.
\begin{lemma}{(Concentration Inequality for Variance Estimation):}
\label{le:le1}
Let $X_1, X_2, \dots, X_n$ be a sequence of i.i.d. random variables bounded in $[l,u]$ with variance $\sigma^2$. Let $\bar{V}(n)= \frac{1}{n-1}\sum^{n}_{i=1}\left(X_i-\frac{1}{n}\sum_{j = 1}^nX_j\right)^2$ be the unbiased estimator of $\sigma^2$. Then, the following concentration inequality holds:
\begin{equation*}
    \mathbb{P}\left(\left|\bar{V}(n) - \sigma^2\right| > \epsilon\right) \leq 2\exp\left(\frac{-2{n}\epsilon^2}{(u-l)^4}\right).
\end{equation*}
\end{lemma}
For random variables supported on $[0, 1]$, concentration inequality for variance is presented in \cite{hou2022almost}. For the case, the domain of the rewards is $[l, u]$, the result follows immediately. Thus, the proof is omitted.
\begin{theorem}
\label{theo:UCB_VV}
For a bandit setting with bounded rewards on $[l=0,u=1]$, the misallocation of \texttt{UCB-VV} is bounded as
\begin{align*}
    \mathcal{M}_{\texttt{VV}}(n) \leq 8 \sum_{i: \sigma_i^2<\sigma_*^2}\frac{\log{n}}{\delta_i^2} + 1 + \frac{\pi^2}{3}.
\end{align*}
\end{theorem}
\begin{IEEEproof}
 See Appendix \ref{App:1}.
\end{IEEEproof}

A key point to note is that while the algorithmic structure follows a \ac{UCB}-like template leveraging confidence bounds to balance exploration and exploitation. However, the statistical analysis is specifically tailored to the variance objective.

\begin{algorithm}[tb]
 \caption{\texttt{UCB-VV}}
 \label{alg:algorithm_vv}
     \textbf{Input}: {$S_{\rm P}, K, n$} \\
     \textbf{Parameter}: $s_i(0)=0$, $\bar{X}_{i}(0)=0$, $\bar{V}_{i}(0)=0$ \vfill
     \begin{algorithmic}[1]
     \FOR{each $t=0, 1, \dots, S_{\rm P}n - 1$}
     \STATE Play arm $i = (t\, \text{mod}\, K) +1$.
     \STATE Update $s_i(t), \bar{X}_{i}(t)$, and $\bar{V}_{i}(t)$.
     \ENDFOR
     \STATE Calculate $B_i(S_{\rm P}n - 1)_{\texttt{VV}}$ $\forall\, i$.
     \FOR{each $t=S_{\rm P}n, S_{\rm P}n+1, \dots, n$}
         \STATE Play arm $i = \argmax\limits_{i \in \{1, 2, \dots, K\}}B_i(t-1)_{\texttt{VV}}$.
         \STATE Update $s_i(t), \bar{X}_{i}(t)$, and $\bar{V}_{i}(t)$.
         \STATE Calculate $B_i(t)_{\texttt{VV}}$.
     \ENDFOR\\
\end{algorithmic}
\end{algorithm}

\subsection{Preliminaries for lower bound}
In this section, we establish the order optimality of the proposed algorithm \texttt{UCB-VV} by deriving the lower bound of the problem, and show it to be matching up to constant factors with the upper bound of $\mathcal{O} \log (n)$. For that, we start with a couple of basic definitions.

\begin{definition}
The ${\rm KL}$ divergence between 2 distributions $\nu$ and $\tilde{\nu}$ is given by
\begin{align*}
    I(\nu,\nu^\prime)= \mathbb{E}_{\nu}\bigg[\log \frac{f_{\nu}(X)}{f_{\tilde{\nu}}(X)}\bigg],
\end{align*}
where $E_{\nu}$ denotes the expectation operator with respect to $\nu$, $f_{\nu}(X)$ is the probability density function of $\nu$ and  $f_{\tilde{\nu}}(X)$ is the probability density function of $\tilde{\nu}$. 
\end{definition}

\begin{assumption}
For any $\theta$, $\lambda$, and $\lambda ^\prime  \in \mathcal{U}$, and for any $\epsilon > 0$, there exists a $\zeta> 0$ such that $0 < \sigma^2(\lambda^\prime)- \sigma^2(\lambda)  < \zeta$ implies $|I(f(.; \theta), f(.; \lambda^\prime))- I(f(.; \theta), f(.;\lambda))| < \epsilon$. This condition ensures that small changes in the variance of distributions result in small changes in their KL-divergences. Thus, implies that the parameter space $\mathcal{U}$ is sufficiently dense to allow fine distinctions between different arms.
\end{assumption}

Since we are interested in the variance of the arms, the rewards can be transformed to have equal means without altering the relative sequence of corresponding variances. As an example, consider a scenario in which we have $2$ sequences $X$ and $Y$ with $r$ \ac{i.i.d} random variables such that $X: X_1, X_2, \dots X_r$ and  $Y: Y_1, Y_2, \dots Y_r$ with the means $\mu_X$ and $\mu_Y$,  and variances $\sigma^2_X$ and $\sigma^2_Y$, respectively. From these, we construct two difference sequences $Z$ and $W$ such that $Z: Z_1=X_2 -X_1, Z_2= X_3 -X_2, \dots Z_{r-1} = X_r-X_{r-1}$ and $W: W_1= Y_2 -Y_1, W_2=Y_3 -Y_2, \dots W_{r-1}= Y_r-Y_{r-1}$. The mean of the resultant sequences is $\mu_Z = \mu_W =0$ but the variances are  $\sigma^2_Z = 2\sigma^2_X$ and $\sigma^2_W = 2\sigma^2_Y$. We see that the variances of the transformed variables are proportional to the variance of the original variables. So, $0 < \sigma^2_Z- \sigma^2_W < \zeta$ implies their distance with $f(.; \theta)$ will also be close, i.e,  $|I(f(.; \theta), f(.; Z))- I(f(.; \theta), f(.;W))| < \epsilon$.

\begin{definition}
A policy $\pi$ is $\alpha$-consistent $0<\alpha<1$, if for one parameter distributions and for all $i\neq *$, $\mathbb{E}[s_i(n)] \leq n^\alpha,$ where $*$ represents the optimal arm.
\end{definition}

Specifically, we assume that the distribution of the arm $i$ is given by $f(.; \theta_i)$ and the distribution model $\mathcal{F} = (f(.;  \theta_1), ..., f(.; \theta_K ))$ can be represented by ${\Theta} = (\theta_1 , ..., \theta_K )$. The parameters $\theta_i$ take values from a set $\mathcal{U}$ satisfying the following regularity conditions (similar
to that in \cite{lai1985asymptotically}).
\begin{assumption}
For all $\theta$ and $\lambda \in \mathcal{U}$, let $X$ be a sub-Gaussian
random variable with distribution $f(.; \theta)$. The random variable
$Y = f(X; \lambda)$ is sub-Gaussian. The assumption of sub-Gaussianity is required for using the Chernoff bound later in the proof, where we invoke this assumption.
\end{assumption}
\begin{lemma}
Let $\Theta$ be the given distribution model, and let
$i$ denote the index of a suboptimal arm under $\Theta$. Under Assumption 1,  number of times arm $i$ is played under $\pi$ for any constant $C_1 < 1-\alpha$,
\begin{align*}
    \lim_{n \rightarrow \infty} \mathbb{P}_\mathcal{F}\bigg[s_i(n) \geq  \frac{C_1 \log n}{I(f_i,f_{*})}\bigg]=1.
\end{align*}
Furthermore, with Assumption 2, there exists $n_o \in \mathbb{N}$ such that
     \begin{align*}
     \mathbb{P}_\mathcal{F}\bigg[s_i(n) \geq  \frac{C_1 \log n}{I(f_i,f_{*})}\bigg]\geq C_2, \quad \textit{for all}\quad  n >n_0
\end{align*}
where constant $0 < C_2 < 1$ is independent of $n$ and $\mathcal{F}$.
\end{lemma}
\begin{IEEEproof}
 See Appendix \ref{App:2}.
\end{IEEEproof}

\subsection{Lower bound on misallocation}
\begin{theorem} Consider the \ac{MAB} problem where the objective is to identify the arm with the highest variance. Let $\pi$ be an $\alpha$-consistent
policy and $\Theta = \left\{ \left\{\theta_i \right\}_{i=1}^K \colon \theta_i \in \mathcal{U}\right\}$ be the distribution model. Under Assumption 1, for any constant
$C_1 < 1-\alpha$, $0< C_2 <1$, with $f_i$ and $f_*$ as the reward distribution of the i-th arm and the optimal arm respectively, the model-specific misallocation satisfies,
\begin{align*}
    \liminf_{n \rightarrow \infty} \frac{\mathcal{M}_{\texttt{VV}}(n)}{\log n} \geq \sum_{i=1,
    i\neq *}^K \frac{C_1C_2}{I(f_i,f_{*})}.
\end{align*}
\end{theorem}
\begin{IEEEproof}
See Appendix \ref{App:3}.
\end{IEEEproof}

From Theorem 1 and Theorem 2, we see \texttt{UCB-VV} achieves an upper bound on the misallocation of $\mathcal{O} \log (n)$ and matches a lower bound of the same order up to constant factors, demonstrating that our proposed algorithm is asymptotically optimal.

\section{Problem Formulation for BAI}
\label{sec:3}
Next, we examine the challenge of \ac{BAI} within a fixed budget. The goal of the learner is to identify and recommend the arm with the highest variance using $n$ samples. Specifically, we propose an algorithm called \texttt{SHVV}, that draws motivation from the foundational framework of sequential halving algorithm introduced by Karnin \textit{et al}~\cite{karnin2013almost}. We explore a scenario where we have a predetermined budget of $n$, aiming to maximize the likelihood of accurate identification of the highest variance arm. The total budget $n$ is divided equally into $\log_2(K)$ elimination rounds. Let $A_r$ be the set of arms in round $r$, which are pulled an equal number of times. After completing a round, we eliminate half of the arms with the lowest estimated variance. The goal of \ac{BAI} setting is to minimize the error probability, i.e., recommending  a sub-optimal arm at the end of the time horizon, given as,
\begin{align*}
   e_n = \mathbb{P}(J_n \neq *).
\end{align*}
Here $J_n$ represents the recommendation. The \texttt{SHVV} algorithm is given in Algorithm \ref{alg:Seq_VV}.

\subsection{Upper bound on error for \ac{BAI}}
In this section, we establish preliminary lemmas to provide the foundation for the subsequent theorem, which is the upper bound on the probability of error for \texttt{SHVV}, specifically focusing on the erroneous elimination of the optimal arm. The analysis hinges on inspecting the empirical third quartile of the surviving arms at each round, which serves as a critical statistic for guiding the elimination process. We first present Lemma~\ref{lemma2:BAI_VV}, which quantifies the likelihood that any suboptimal arm $i \in A_r$ outperforms the optimal arm.
\begin{lemma}
\label{lemma2:BAI_VV}
Assuming that the best arm is not eliminated prior to round $r$. Then for any arm $i \in A_r$
\begin{align*}
    \mathbb{P}\left(\bar{V}_i(t_r)>\bar{V}_* (t_r) \right)\leq \exp\left(-\frac{(t_r-1)^2 \delta_i^2}{2t_r}\right)
\end{align*} 
where $t_r= \frac{n}{|A_r|\log_2(K)}$.
\end{lemma}
\begin{IEEEproof}
This probability is evaluated as
\begin{align*}
    &\mathbb{P}(\bar{V}_i(t_r)>\bar{V}_*(t_r)) = \mathbb{P}\bigg(\frac{1}{t_r-1 }\sum_{q=1} ^{t_r} \left(X_i(q)-\bar{X}_i(t_r)\right)^2 - \\
    & \frac{1}{t_r-1} \sum_{q=1} ^{t_r} \left(X_*(q)-\bar{X}_*(t_r)\right)^2 -(\sigma^2_i-\sigma^2_*) > \delta_i \bigg)    \nonumber \\
    &\leq \exp\left(-\frac{(t_r-1)^2 \delta_i^2}{2t_r}\right).
\end{align*}
The last inequality follows from Lemma (\ref{le:le1}).
\end{IEEEproof}

Building upon this foundation, Lemma~\ref{lemma3:BAI_VV} provides a complementary result by quantifying the probability that the optimal arm is erroneously eliminated during round $r$.
\begin{lemma}
\label{lemma3:BAI_VV}
The probability that the best arm is eliminated in round $r$ is at most
\begin{align*}
    3 \exp \left(\frac{-\left(n-4i_r\log_2(K)\right)^2\delta_{i_r}^2}{ 8 n i_r \log_2(K)}\right)
\end{align*}
where $i_r = \frac{K}{2^{r+2}}$,   $|A_r| = 4 i_r$.
\end{lemma}
\begin{IEEEproof}
See Appendix \ref{App:4}.
\end{IEEEproof}

Leveraging the results of Lemmas~\ref{lemma2:BAI_VV} and~\ref{lemma3:BAI_VV}, we now state and prove the main theorem, which establishes an upper bound on the overall probability of error.

\begin{theorem}
\label{theo:BAI_VV}
The probability of error is upper bounded as
\begin{align*}
    e_n \leq   3 \log_2(K) \exp\left(\frac{-\left(n - K \log_2{K} \right)^2} {8n\log_2(K) H_2}\right)
\end{align*}
where $H_2 := max_{i\neq *}\frac{i}{\delta_i^2}$.
\end{theorem}
\begin{IEEEproof}
Using Lemma \ref{lemma3:BAI_VV} and the union bound, the best arm is eliminated in one of $\log_2(K)$ phases with probability at most 
\begin{align*}
    &3 \sum_{r=1} ^{\log_2(K)} \exp \left(-\frac{\left(n-4i_r\log_2(K)\right)^2\delta_{i_r}^2}{ 8 n i_r \log_2(K)}\right)\\
    & \leq 3 \log_2(K) \exp\Bigg(-\frac{(n-4i_r\log_2(K))^2}{8 n\log_2(K)} \times  \frac{1}{\max_i i\delta_{i}^{-2}}\Bigg)\\
    & \leq 3 \log_2(K) \exp\left(\frac{-\left(n - K \log_2{K} \right)^2}{8 n\log_2(K)  H_2}\right).
\end{align*}
\end{IEEEproof}

\begin{algorithm}[t]
\caption{\texttt{SHVV}}
\label{alg:Seq_VV}
    \textbf{Input}: {$ K,n $} \\
    \textbf{Parameter}:  $\bar{V}_{i}=0$ \vfill
    \textbf{Initialize}: $A_1 \leftarrow K$, $ r \leftarrow 1$ 
    
    \begin{algorithmic}[1]
    \FOR{ $ r = 0$ to $ \lceil\log_2 K\rceil -1 $ } 
    \STATE Play arm $a_i \in A_r$ for $t_r= \frac{n}{|A_r|\log_2(K)}$
    \STATE Update  $\bar{V}_{i}$.
    \STATE Calculate the reward
    \STATE Let $A_{r+1}$ be the set of $\lceil\frac{A_r}{2}\rceil$ arms in $A_r$ with largest empirical average
     \ENDFOR\\
     \textbf{output} arm in $A_{\lceil\log_2 K\rceil}$
   \end{algorithmic}
\end{algorithm}

\subsection{Lower bound on error for \ac{BAI}}
To establish the lower bound, we examine $K$ bandit instances, each consisting of $K$ arms but each characterized by a distinct optimal arm. The learner is aware that the bandit environment it encounters must correspond to one of these $K$ bandit settings. The lower bound is derived by proving that the learner, however good it is, will nevertheless make a mistake in identifying the bandit instance for a set of observations. Let $G^B$ be the setting where $B \in \{1, \dots K\}$ is optimal. Accordingly,
\begin{align*}
    G^1 ;= \nu_k \sim U [l_k, u_k] \quad \forall k \in \{1, \dots K\},
\end{align*}
for $0\leq l_1 \leq l_k \leq u_k \leq u_1 \leq 1$, i.e., the first arm has the highest variance.
Let there be another bandit setting $G^{k}$:
\begin{align*}
    G^{k} &=
    \begin{cases}
        \nu_k = \nu_i & i \neq k \\
        \nu_k \sim U \left[\tilde{l}_{k}, \tilde{u}_k \right] & i = k
    \end{cases}
\end{align*}
where $0 \leq \tilde{l}_{k}\leq l_k \leq u_k \leq \tilde{u}_{k} \leq 1$. Note, ${\tilde{\sigma}^2_{k}} > \sigma_1^2 > \sigma^2_k$.
\begin{lemma}
\label{lemma4:BAI_VV}
Let $Z_1, Z_2, \dots, Z_n$ be a sequence of i.i.d. random variables such that $|Z_s| \leq M$ for all $s$, then for $\epsilon>0$, by Hoeffding's inequality  we have 
\begin{align*}
    \mathbb{P}\left(\left|\frac{1}{n} \sum_{s=1}^n Z_s - \mathbb{E}[Z] \right| \geq \epsilon  \right) \leq 2\exp\left(\frac{-2{n}\epsilon^2}{M^2}\right).
\end{align*}
\end{lemma}
\begin{theorem}
For any bandit strategy that returns an arm $J_n$ at time $n$, for constant $M > 0$, it holds that:
\begin{align*}
    &\max_{1 \leq k \leq K} \mathbb{P}_{k}\left(J_n\neq k \right) \geq \\
    &\hspace*{2cm} \frac{1}{6} \exp\left(-3 \frac{n}{H_\sigma(1)  \sigma_{\nu}^2}  - \frac{M}{\sqrt{2}}\sqrt{n \log(6nK)}\right)
\end{align*}
where $\mathbb{P}_{k}(J_n \neq k)$ is the probability of error under problem  $G^k$,  $H(k) = \sum_{1 \leq i \leq K, i \neq k} \delta_i^{-2}$ measures the difficulty of identifying the best arm in problem $G^k$, $H_\sigma(1) =  H(1) \delta_k$.
\end{theorem}
\begin{IEEEproof}
See Appendix \ref{App:5}.
\end{IEEEproof}

\section{Sub-Gaussian Distributions and Sharpe Ratio}
\label{sec:4}
In this section, we extend our framework from bounded distributions to sub-Gaussian distributions. Specifically, we first derive a concentration inequality for the sample variance and standard deviation of sub-Gaussian random variables. Building on this, we extend our analysis to establish novel concentration bounds for the \ac{SR}.
\begin{theorem}
\label{th:theo_5}
Let $X_1, X_2, \dots, X_n$ be i.i.d. sub-Gaussian random variables with mean 
$\mu$, variance $\sigma^2$, and sub-Gaussian parameter $v^2$, satisfying
\begin{align*}
\mathbb{E}\!\left[\exp\!\left(\lambda (X_i - \mu)\right)\right] 
\le \exp\!\left(\frac{\lambda^2 v^2}{2}\right), \qquad \forall \lambda\in\mathbb{R}
\end{align*}
Then, there exists a constant $C_V>0$ such that for any $\epsilon>0$,
\begin{align*}
\mathbb{P}\left(|\bar{V}(n)-\sigma^2|>\epsilon\right)
\le 
4 \exp\!\left(
- C_V n \min\!\left(\frac{\epsilon^2}{v^4}, \frac{\epsilon}{v^2}\right)
\!\right)
\end{align*}
\end{theorem}
\begin{IEEEproof}
See Appendix \ref{App:6}.
\end{IEEEproof}
\begin{corollary}[Concentration of the Empirical Standard Deviation]
Under the assumptions of Theorem~\ref{th:theo_5},  
we define the empirical standard deviation $\sqrt{\bar{V}(n)}$.  
Then for any $\epsilon>0$, there exists a constant $C>0$ such that
\begin{align*}
\mathbb{P}\!\left(\left|\sqrt{\bar{V}(n)}-\sigma\right| > \epsilon \right)
\le 4 \exp\!\left(
- C n \min\!\left(\frac{\epsilon^2}{v^4}, \frac{\epsilon}{v^2}\right)
\right).
\end{align*}
\label{Col:1}
\end{corollary}

\begin{figure*}[t]
    \centering
    \subfloat[]
    {\includegraphics[width=0.3\linewidth]{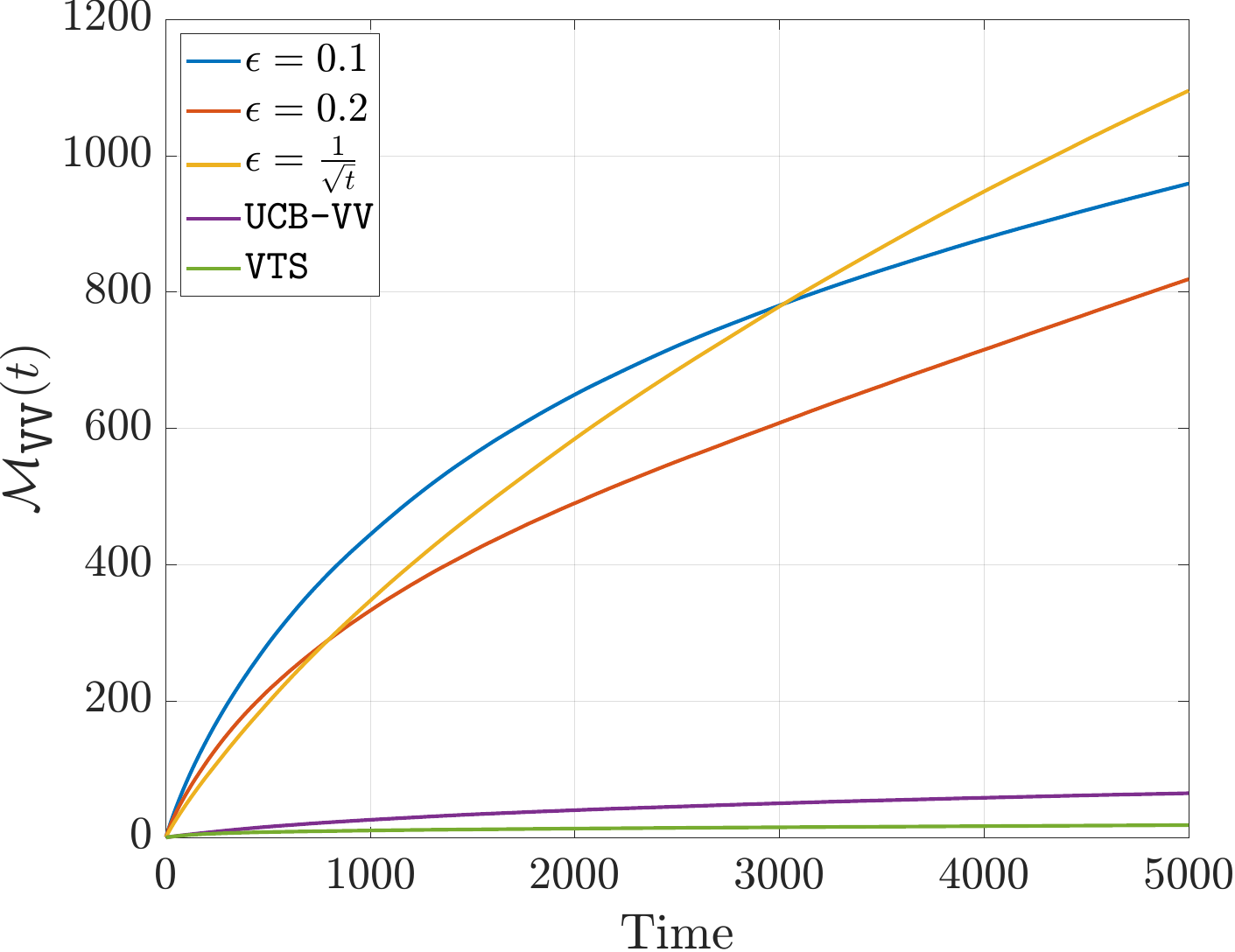}
    \label{fig:fig3_a}}
    \hfil
    \subfloat[]
    {\includegraphics[width=0.3\linewidth]{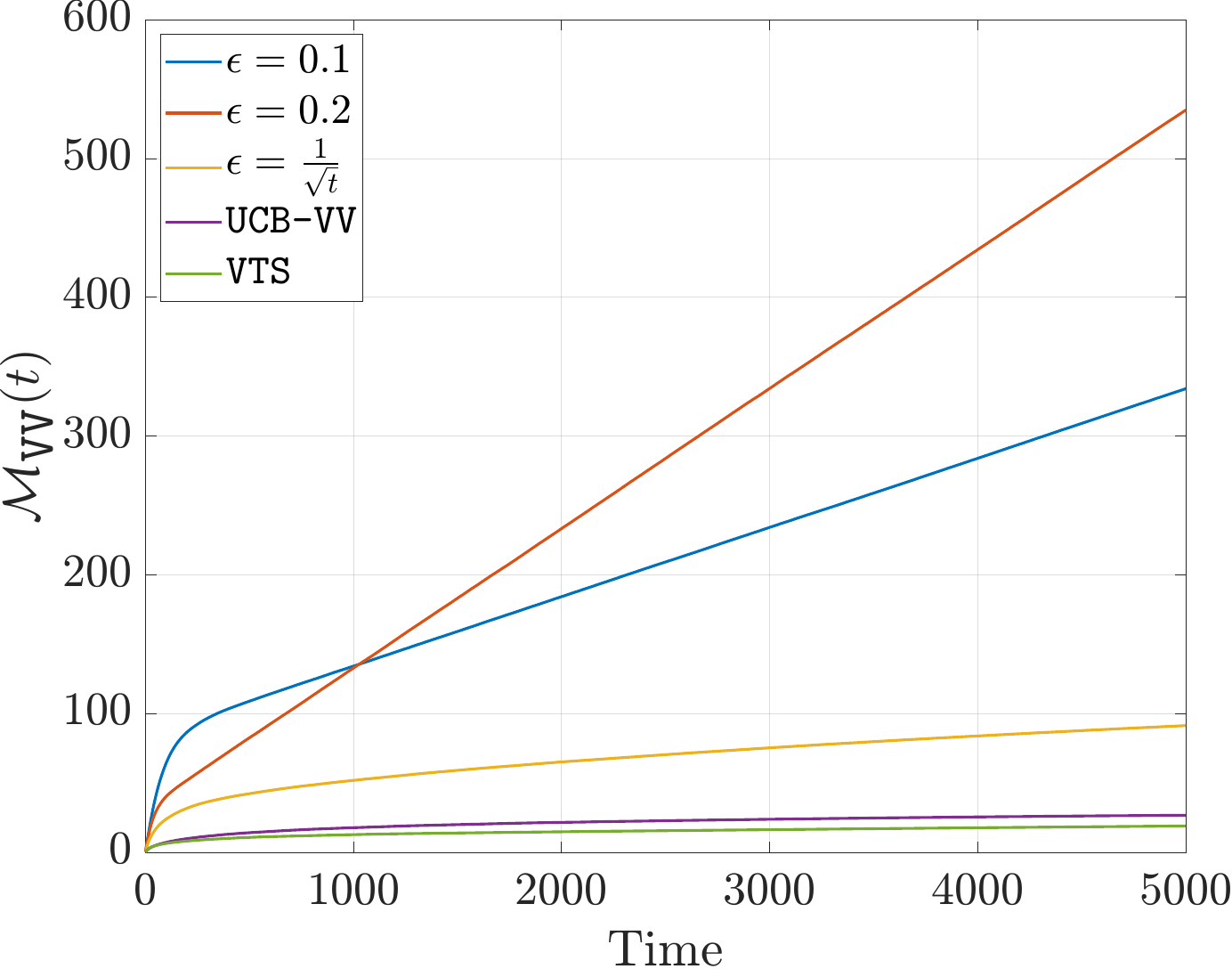}
    \label{fig:fig3_b}} 
    \hfil
    \subfloat[]
    {\includegraphics[width=0.3\linewidth]{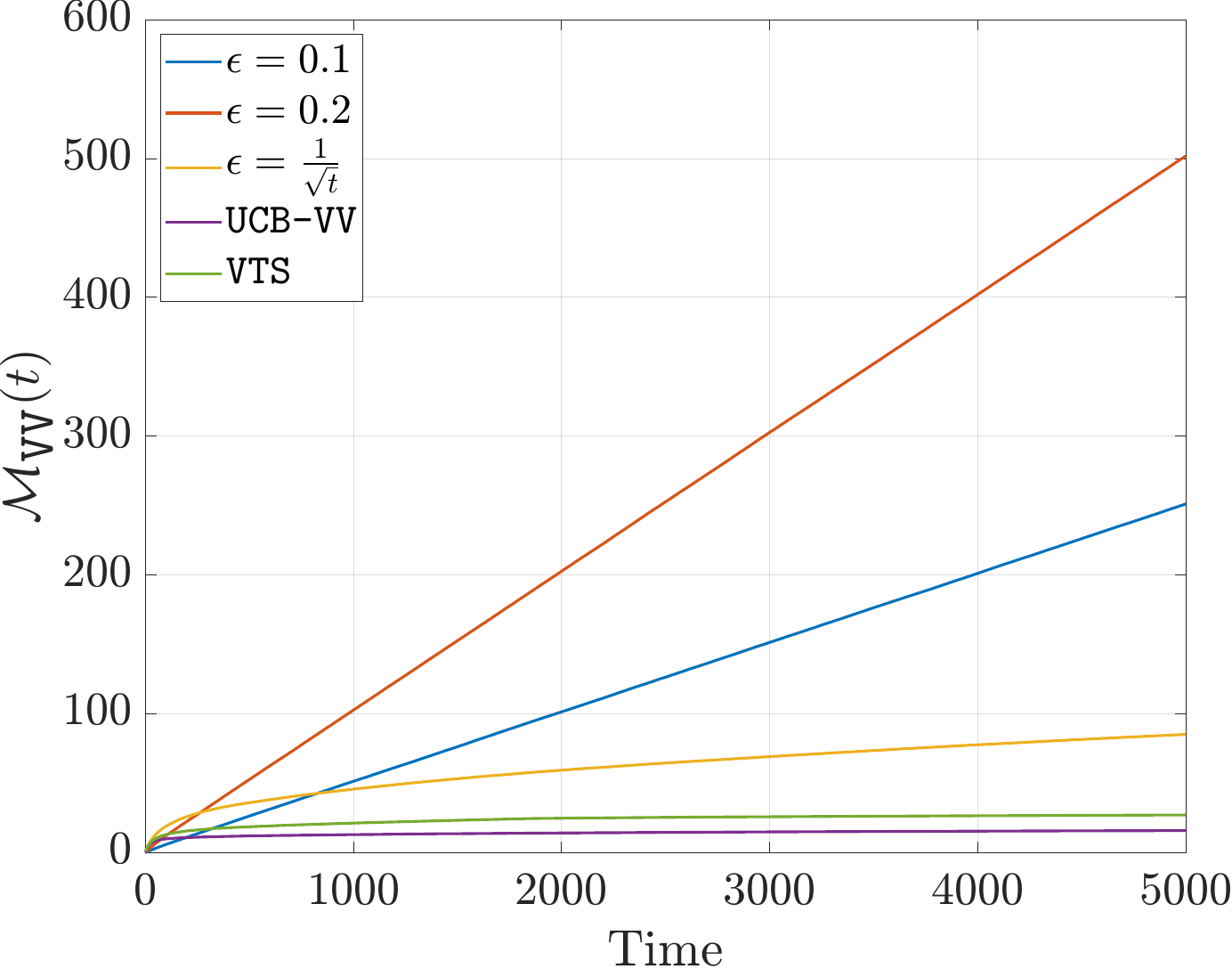}
    \label{fig:fig3_c}} 
    \caption{Misallocation v/s time for \texttt{UCB-VV}, \texttt{$\epsilon$-Greedy} and \texttt{VTS} for (a) $\delta = 0.1$, (b) $\delta = 0.5$, and (c) $\delta = 1$.}
    \label{fig:fig4}
\end{figure*}

\subsection{Concentration inequality for Sharpe Ratio}
The \ac{SR}, defined as the ratio of expected reward to its standard deviation, is a standard metric for evaluating performance relative to risk in sequential decision-making frameworks. While our earlier research on bounded random variables derived path-dependent concentration inequalities for the \ac{SR} \cite{khurshid2025optimizing}, real-world scenarios often involve unbounded rewards. Extending results to the sub-Gaussian distributions not only broadens their applicability but also streamlines the analysis by removing dependencies of trajectory-specific bounds tied to bounded variables. Here, we establish a new concentration bound for the sample \ac{SR} for sub-Gaussian random variables.
\begin{theorem}
\label{theo: Th6}
Let $ X_1, X_2, \dots, X_n $ be i.i.d. sub-Gaussian random variables with mean $ \mu = \mathbb{E}[X_i] $, standard deviation $ \sigma = \sqrt{\mathbb{E}[(X_i - \mu)^2]} $, and sub-Gaussian parameter $ v^2 $. If the true \ac{SR} is $S = \frac{\mu}{\sigma}$ and empirical \ac{SR} is $\bar{S}(n) = \frac{\bar{\mu}(n)}{\sqrt{\bar{V}(n)}}$, there exists a constant $ c > 0 $ depending on $ \mu, \sigma^2, v^2 $ such that:
\begin{align*}
\mathbb{P}\left(|\bar{S}(n) - S| \geq \eta\right) \leq 4 \exp\left(-c n \min\left(\eta^2, \eta\right)\right).
\end{align*}
\end{theorem}
\begin{IEEEproof}
See Appendix \ref{App:7}.
\end{IEEEproof}

\subsection{UCB-Sharpe algorithm and its regret}
Leveraging the concentration inequality above, we propose the \texttt{UCB-Sharpe} algorithm that selects arms with the highest \ac{SR}, balancing empirical performance with uncertainty in estimation. Let each arm $ i \in \{1, \dots, K\} $ have unknown mean $ \mu_i $, standard deviation $ \sigma_i $, and true \ac{SR} $ S_i = \mu_i / \sigma_i $. The UCB-Sharpe index at time $t$ is $\text{Index}_i(t) = \bar {S}_i(t)+ \beta_i(t)$, where $\bar {S}_i(t) = \bar{\mu}_i/ \bar{\sigma}_i $ is the estimated \ac{SR} and $\beta_i(t)$ is the confidence bonus derived from the concentration inequality for the sample \ac{SR} and is equal to $ \beta_i(t) = \sqrt{\frac{1}{C s_i(t)} \log(4t^2)}$. The steps are presented in Algorithm \ref{alg:algorithm_UCB-SR}. .The constant $C$ required in the algorithm is a lower bound on the $c$ (from Theorem 6) for each of the $K$ arms.

\begin{algorithm}[tb]
 \caption{\texttt{UCB-Sharpe}}
 \label{alg:algorithm_UCB-SR}
     \textbf{Input}: {$K, n, c, v^2$} \\
     \textbf{Initialize}: Pull each arm once \vfill
     \begin{algorithmic}[1]
     \FOR {$t=K+1$ to  $T$} 
       \FOR{each arm $i=1, \dots, K$} 
       \STATE Compute $\bar{\mu}_i(t)$ and $\sqrt{\bar{V}(t)}$
       \STATE Compute estimated \ac{SR} $ \bar {S}_i(t) = \frac{ 
              \bar{\mu}_i}{\bar{\sigma}_i} $
       \STATE Compute confidence bonus $ \beta_i(t) = \sqrt{\frac{1}{C s_i(t)} \log(4t^2)}$
       \STATE Set index: $\text{Z}_i(t) = \bar {S}_i(t)+ \beta_i(t)$       \ENDFOR
        \STATE Pull arm $I(t)= \argmax_i \text{Index}_i(t)$
     \ENDFOR\\

\end{algorithmic}
\end{algorithm}

\begin{algorithm}[tb]
\caption{Variance Thompson Sampling - \texttt{VTS}~\cite{zhu2020thompson}}
\label{alg:alg_VTS}
\textbf{Input:} $x_{i,\max}(0) = 0, s_{i}(0) =0$ \vfill
\begin{algorithmic}[1]
    \STATE \hspace{0.5cm} \textbf{For} each $t = 1, 2, \dots, K$ \textbf{do}
    \STATE \hspace{0.9cm} Play arm $t$
    \STATE \hspace{0.9cm} Update $(x_{i,\max}(t), s_{t}(t-1))$
    \STATE \hspace{0.6cm}\textbf{end For}
    \STATE \hspace{0.5cm} \textbf{For} each $t = K+1, K+2, . . . , n$ \textbf{do}
    \STATE \hspace{0.9cm} Sample $\tilde{\theta}_{i,t} \sim \mathrm{Pareto}\big(x_{i,\max,t-1},\, n_{i,t-1}\big)$
    \STATE \hspace{0.9cm} Play arm $i(t) = \arg\max_{i\in[K]} \tilde{\theta}_{i,t}^{2} / 12$ and observe reward $X_{i(t),t}$
    \STATE \hspace{0.9cm} Update $(x_{i(t),\max}(t), s_{i(t)}(t-1))$
    \STATE \hspace{0.6cm}\textbf{end For}
\end{algorithmic}
\end{algorithm}

Let $ S_* = \max_i S_i $ be the optimal \ac{SR}, then the sub-optimality gap for arm $i$ is $\Delta_i = S_* - S_i $. The cumulative regret at time $n$ is 
\begin{align}
\text{R}_{\texttt{SR}}(n) = \sum_{i: \Delta_i > 0} \Delta_i \cdot \mathbb{E}[s_i(n)],
\label{eq:reg-SR}
\end{align}
where $ s_i(n) $ is the number of times arm $i$ is pulled by time $n$.
\begin{theorem}
If \texttt{UCB-SR} is run for $K$ arms having sub-Gaussian reward distributions with parameter $v^2$, then its expected regret after $n$ of plays is at most
 \begin{align*}
     \text{R}_{\texttt{SR}}(n) \leq \sum_{i: \Delta_i > 0} \left( \frac{9}{c \Delta_i} \log n + \mathcal{O}(1) \right).
   \end{align*}
\end{theorem}
\begin{IEEEproof}
See Appendix \ref{App:8}.
\end{IEEEproof}
The regret scales logarithmically with $n$, similar to classical \ac{UCB} algorithms, depending on the parameter $c > 0$ which is a function of sub-Gaussian parameter $v^2$, $\mu$ and $\sigma^2$.

\section{Numerical results}
\label{sec:5}
\subsection{Performance for \texttt{UCB-VV}}
In Fig.~\ref{fig:fig4}, we compare the performance of our proposed algorithm \texttt{UCB-VV} with existing algorithms like \texttt{VTS} and \texttt{$\epsilon$-greedy}. \texttt{VTS}~\cite{zhu2020thompson} extends classical \ac{TS} to risk-aware decision making by incorporating uncertainty in variance rather than relying solely on mean rewards. While \texttt{VTS}~\cite{zhu2020thompson} primarily considers Gaussian and Bernoulli reward models equipped with appropriate conjugate priors, our setting fundamentally differs in that, as rewards are bounded. For our numerical investigations, we model the reward of the $i$th arm as $X_i \sim \mathrm{Uniform}(0,\theta_i)$, where the unknown parameter $\theta_i>0$ fully characterizes both mean and variance. Consequently, variance maximization is equivalent to the estimation and maximization of the scale parameter $\theta_i$. We adopt a non-informative prior $p(\theta_i )\propto 1/\theta_i$, which is commonly used for scale parameters. Given $s_i(t)$ observations from arm $i$ up to time $t$, let $x_{i,\max}(t)=\max \left\{X_{i,1},\ldots,X_{i,s_i(t)} \right\}$ denote the sample maximum. Under the above prior, the posterior distribution of $\theta_i$ admits the closed-form Pareto density which exhibits a polynomially decaying heavy tail. In our implementation, \ac{TS} for variance maximization proceeds by independently drawing $\tilde{\theta}i(t) \sim \mathrm{Pareto}\big(x_{i,\max}(t), s_i(t)\big)$ for each arm and selecting the arm that maximizes the sampled variance $\tilde{\theta}_i(t)^2/12$. The posterior parameters admit particularly simple online updates. After observing a new reward $X_{i,t}$ from arm $i$, the sufficient statistics are updated as
\begin{align*}
    s_i(t + 1) = s_i(t) + 1, \,\, x_{i, \max}(t + 1) = \max \left\{x_{i,\max}(t), X_{i,t} \right\}.
\end{align*}
These two quantities fully characterize the posterior at every time step, making the proposed \texttt{VTS} implementation computationally lightweight.

We run all algorithms for $2$ arms and a time horizon of $n =5000$. In Fig.~\ref{fig:fig3_a},~\ref{fig:fig3_b} and~\ref{fig:fig3_c} we test our algorithm for both fixed values of ($\epsilon = 0.1, 0.2$) and a decaying schedule ($\epsilon_t = 1/\sqrt{t}$). For \texttt{UCB-VV}, we use the theoretically derived confidence interval i.e, $C_{t,s_i} = \sqrt{\frac{2 \log t}{s_i}}$. For $\delta_i = 0.1$, we see \texttt{UCB-VV} outperforms \texttt{$\epsilon$-greedy}, but the misallocation of \texttt{VTS} is the lowest. In Fig. \ref{fig:fig3_b}, for $\delta_i = 0.2$, we see the misallocation decreases for all algorithms as the distinction between optimal and sub-optimal arms gets easier. In case of $\delta_i = 0.1$, we observe the misallocation for $\epsilon = 0.1$ is larger than for $\epsilon = 0.2$. This is because a larger $\epsilon$ implies larger exploration and thus the algorithm is able to pull each arm a greater number of times and can gather more data about all the arms, before converging on the optimal arm. But excessive exploration can lead to higher misallocation, as we can see in the case of $\delta_i = 0.5$. However, the drop in misallocation is larger for \texttt{UCB-VV} and \texttt{VTS}, and the gap between the misallocation is also smaller. When the sub-optimal variance gap is moderate i.e. $\delta = 0.1, 0.5$, the problem is intrinsically difficult and requires sustained exploration of both arms. In this regime, the randomized exploration of \texttt{VTS} provides a more balanced trade-off between exploration and exploitation, while \texttt{UCB-VV} tends to be relatively conservative. As a result, \texttt{VTS} exhibits superior finite-time performance.

In contrast, when the sub-optimal gap is large i.e. $\delta = 1$, as shown in Fig.~\ref{fig:fig3_c}, the advantage shifts to \texttt{UCB-VV}. In this regime, the deterministic confidence bounds in \texttt{UCB-VV} separate the arms quickly, causing the algorithm to commit to the optimal arm after a short transient phase. The posterior of the uniform distribution under a non-informative prior is Pareto with heavy polynomial tails. Consequently, the \texttt{VTS} occasionally draws extremely large values for the suboptimal arm, even when it has already been sampled many times. These rare but persistent overestimates induce unnecessary exploration of the inferior arm. Although these events become less frequent over time, their persistence over a finite horizon is sufficient to give \texttt{UCB-VV} a clear performance advantage in the large-gap regime.

While our proposed algorithm is optimal and offers theoretical guarantees, the superior performance of \texttt{VTS} highlights the need for a theoretical investigation of \texttt{VTS}. Thus providing the direction for future research, which could further enhance the applicability and impact of \texttt{VTS}.
\begin{figure*}[t]
    \centering
    \subfloat[]
    {\includegraphics[width=0.25\linewidth]{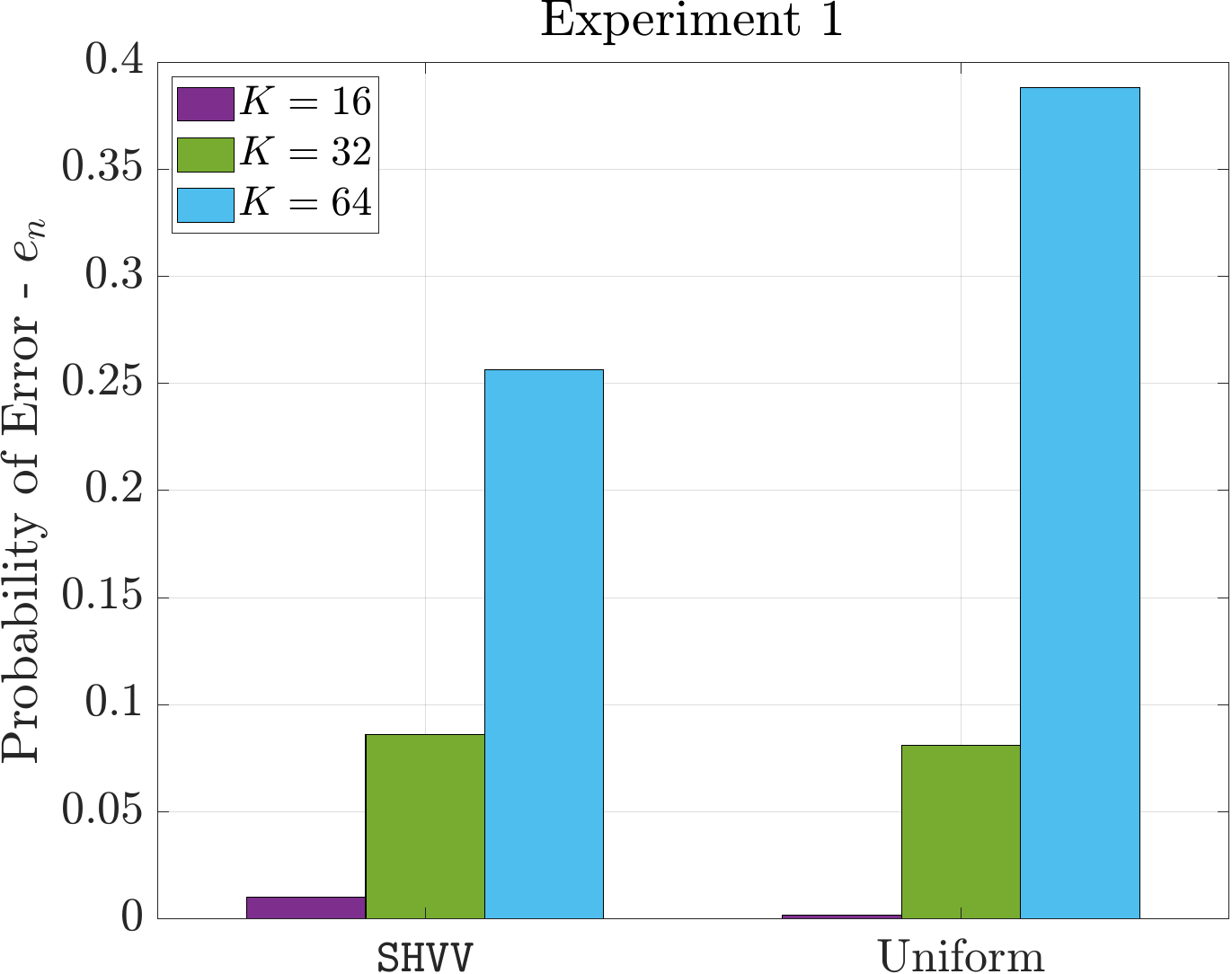}
    \label{fig:fig2_a}}
    \hfil
    \subfloat[]
    {\includegraphics[width=0.25\linewidth]{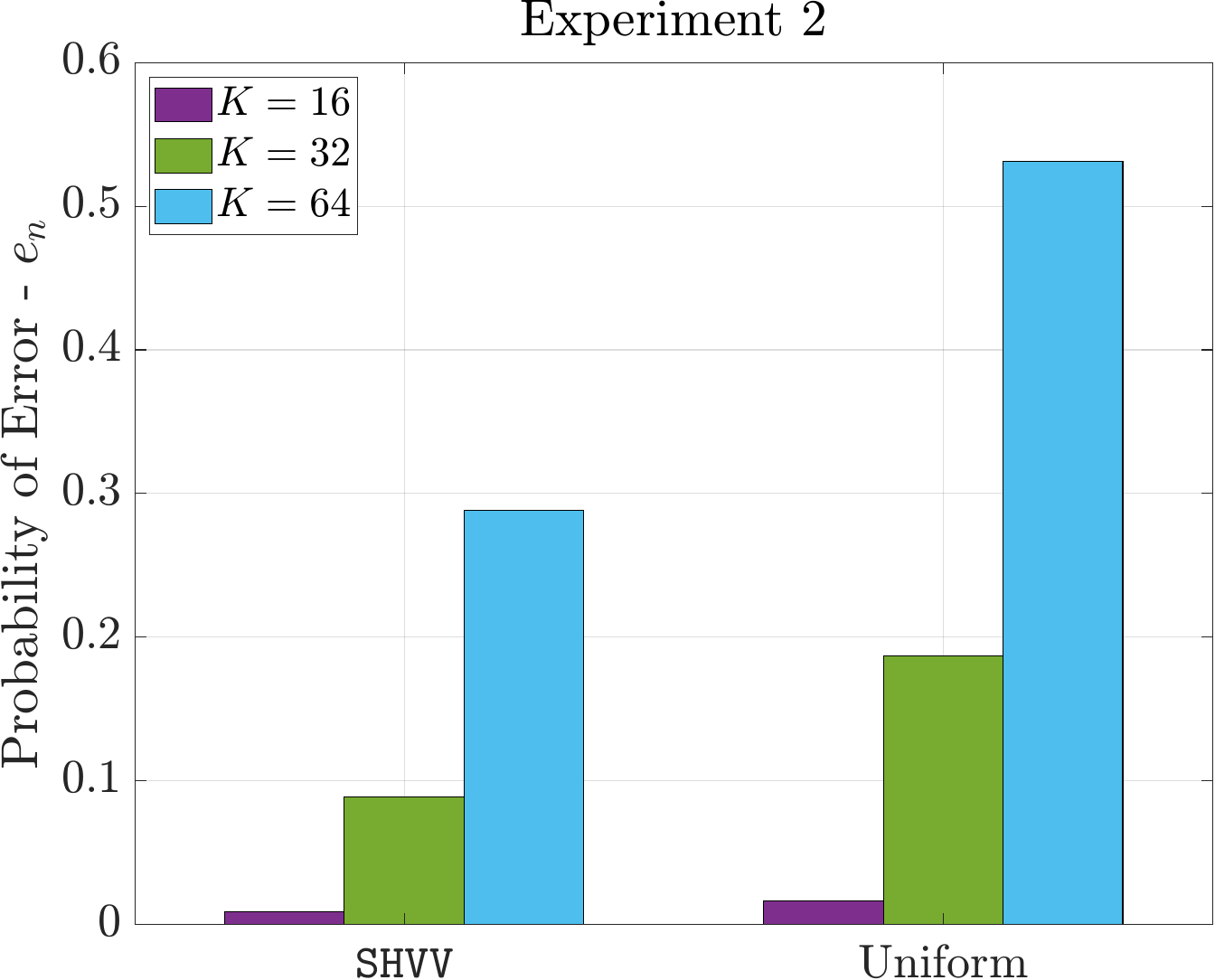}
    \label{fig:fig2_b}}
    \hfil
    \subfloat[]
    {\includegraphics[width=0.25\linewidth]{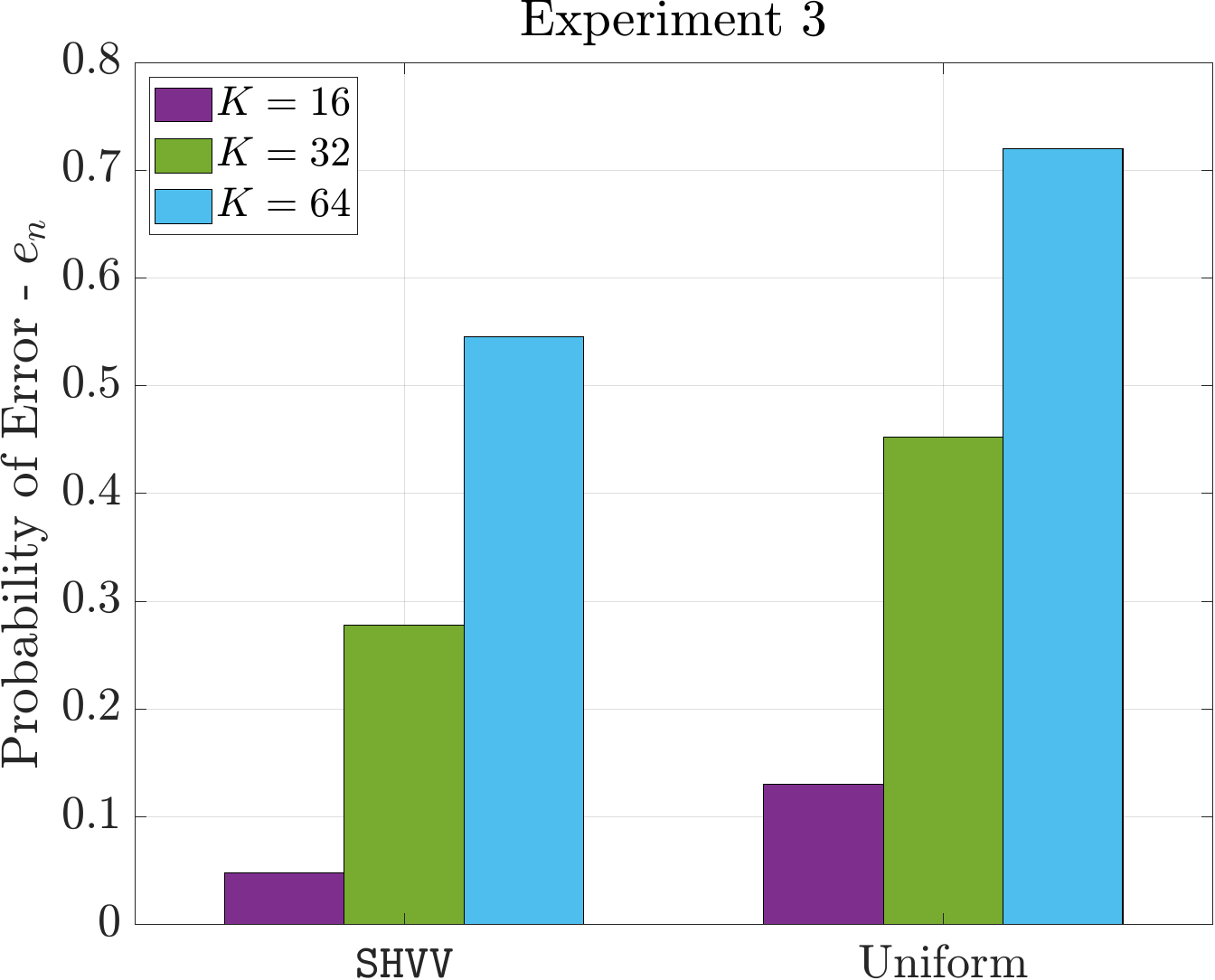}
    \label{fig:fig2_c}} 
    \hfil
    \subfloat[]
    {\includegraphics[width=0.25\linewidth]{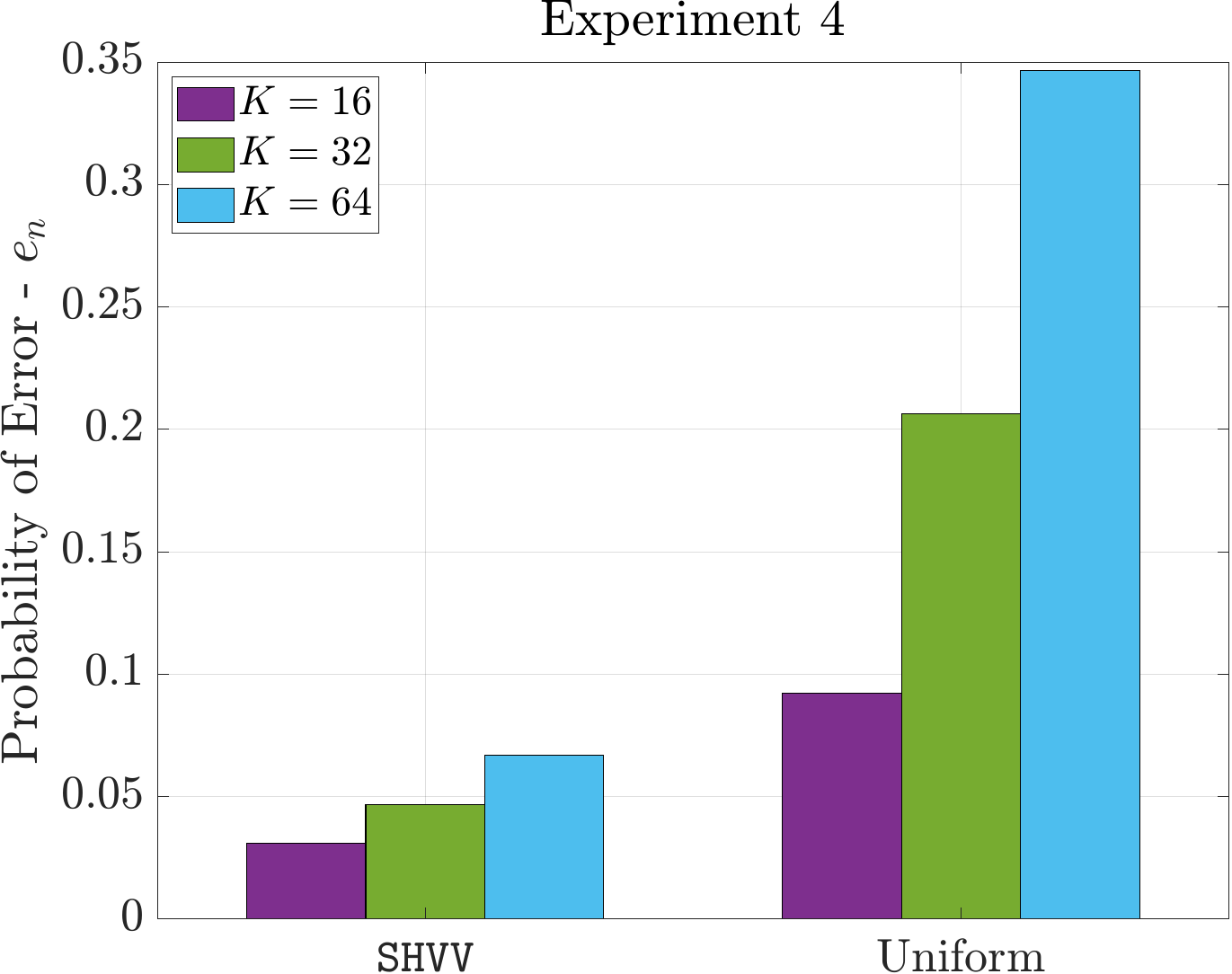}
    \label{fig:fig2_d}}
    \hfil
    \subfloat[]
     {\includegraphics[width=0.25\linewidth]{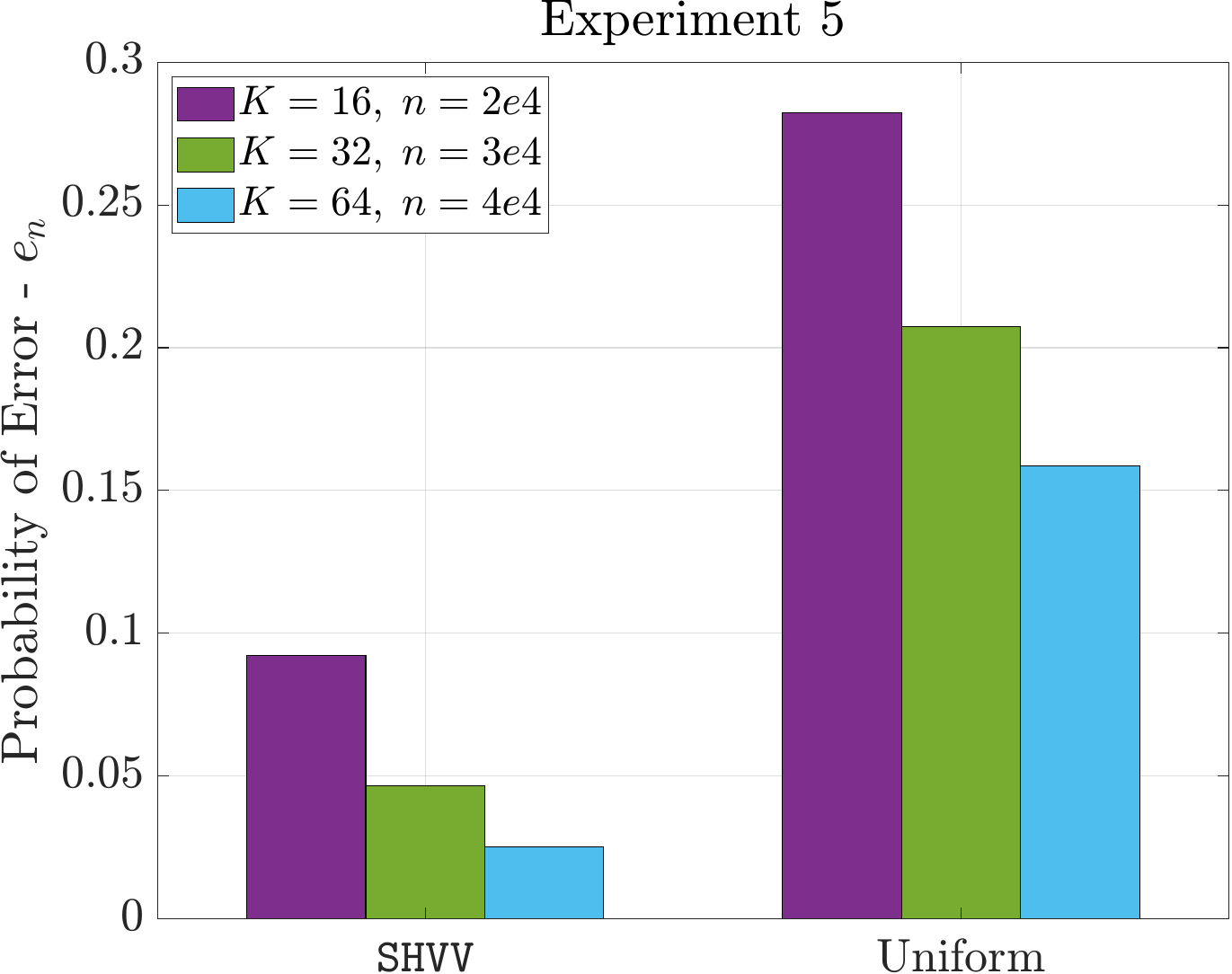}
    \label{fig:fig2_e}} 
    \hfil
    \subfloat[]
    {\includegraphics[width=0.25\linewidth]{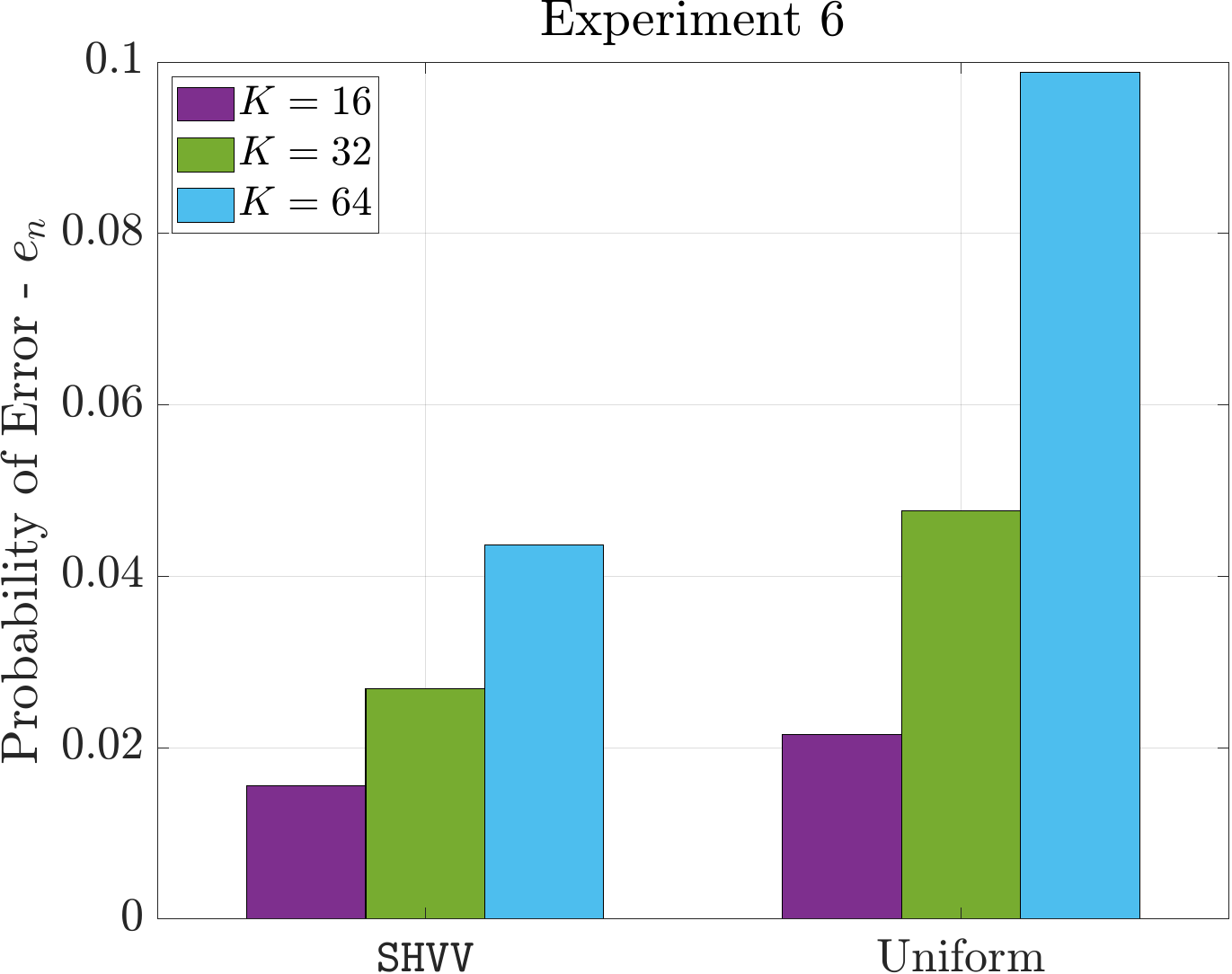}
    \label{fig:fig2_f}}
    \caption{Error probability $e_n$ of $\texttt{SHVV}$ and uniform sampling algorithm for 6 experimental setups defined.}
    \label{fig:fig3}
\end{figure*}

\subsection{Performance for \texttt{SHVV}}
In Fig. \ref{fig:fig3}, we see the performance of \texttt{SHVV} for a fixed budget of $n$ for 6 setups. Each setup presents a unique scenario, as the sub-optimality gaps follow geometric progression and arithmetic progression or can be equal within a group of sub-optimal arms. We maintain the $\sigma_1^2= \frac{1}{12}$ as our optimal arm for all cases.
\begin{enumerate}
     \item \textbf{Experiment - 1:} One group of sub-optimal arms: for $K=16$, $\sigma^2_{2:16} = \frac{1}{15}$, for $K=32$, $\sigma^2_{2:32} = \frac{1}{15}$, and for $K=64$, $\sigma^2_{2:64} =  \frac{1}{15}$.
     \item \textbf{Experiment - 2:} Two groups of sub-optimal arms: for $K=16$, $\left[\sigma^2_{2:6} = \frac{1}{14},\; \sigma^2_{7:16} = \frac{1}{17}\right]$, for $K=32$, $\left[\sigma^2_{2:14} = \frac{1}{14},\; \sigma^2_{15:32} = \frac{1}{17}\right]$ and for $K=64$, $\left[\sigma^2_{2:30} = \frac{1}{14},\; \sigma^2_{31:64} = \frac{1}{17}\right]$.
     \item \textbf{Experiment - 3:} Arithmetic progression: $\sigma_i^2 = \frac{1}{13} - 0.0021(i-2)$, $i \in \{2,\dots,K\}$.
     \item \textbf{Experiment - 4:} Geometric progression: $\sigma_i^2 = \left(\frac{1}{12}\right)0.98^i$, $i \in \{2,\dots,K\}$.
     \item \textbf{Experiment - 5:} Random generation: for each realization, the lower and upper limits are randomly generated with $l,\, u \in [0, 1]$ such that $l < u$.
\end{enumerate}

We observe that as the number of arms increases, $e_n$ increases. We see in Fig. \ref{fig:fig2_a},~\ref{fig:fig2_b}, \ref{fig:fig2_c}, \ref{fig:fig2_d} and \ref{fig:fig2_f}, an error corresponding to the detection of optimal arm increases as $K$ jumps from $16-32$ and $32-64$ for a fixed budget of $2000$. The trend is similar for uniform sampling, although \texttt{SHVV} has the lowest error. This implies that as arms grow, the budget required is greater than provided $n$, for the algorithm to perform optimally with growing arms. 

Also, we see in Fig. \ref{fig:fig2_e}, for \texttt{SHVV}, as the number of arms increases, $e_n$ decreases as the budget allocated is also increased proportionate to the increase in arms. The error of \texttt{SHVV} is the least compared to uniform sampling through all experiments.

\section{Case study: call option trading}
\label{sec:6}
Options are a widely used financial tool that investors employ to speculate on market movements or hedge against potential risks. The potential profits for option buyers are limitless, but their losses are capped, which is why they pay a premium. Conversely, option writers face potentially unlimited losses, while their profits are restricted, which is why they receive a premium. Call options are a fundamental derivative instrument in finance, granting the holder the right, but not the obligation, to purchase an underlying asset at the strike price, within a specified time period \cite{merton1978returns}. The buyer expects stock prices to rise, while the writer anticipates a decline. The call option payoff is given as $\max(0, S_t - R) $,
where $S_t$ is the stock price of the underlying asset at time $t$, $R$ is the strike price, i.e., the predetermined price at which the buyer can purchase the asset. The profit for the buyer is calculated by subtracting the premium $P$ from the payoff. The premium paid is a cost of purchasing the call option.

\subsection{Relation Between Variance and Call Option Payoff}
In this paper, the primary theoretical objective is to study variance identification as a standalone online learning problem, in which the variance of each arm is treated as the reward signal. Accordingly, in the misallocation setting, the learner aims to identify the arm with the highest variance under bandit feedback. Although this departs from the classical mean-reward formulation, it remains a valid stochastic bandit setting: the learner sequentially interacts with unknown stochastic arms, observes noisy rewards, and seeks to minimize misallocation with respect to an unknown optimal arm. The call option trading example serves as a motivating application in which variance acts as a meaningful proxy for expected payoff due to the convex structure of the option payoff.

For a European call option, the payoff $f(S_T) = \max(S_T - K, 0)$ is a convex function of the terminal price $S_T$. Higher volatility increases the probability mass in the upper tail of the asset-price distribution, while the downside is capped at zero. As a result, the expected payoff of a call option increases with volatility. This fundamental property is well established in classical option pricing theory. In the Black-Scholes framework~\cite{black1973pricing}, the call price increases monotonically with the volatility parameter. Merton~\cite{merton1978returns} further notes that a call option written on a more volatile stock is more valuable than one on a less volatile stock. A general economic justification is provided by the theory of mean-preserving spreads introduced by Rothschild and Stiglitz~\cite{rothschild1971increasing}. For random variables $X$ and $Y$, $X \le_r Y \implies Y \overset{d}{=} X + Z, \mathbb{E}\left[Z \mid X\right] = 0$, meaning that $Y$ is riskier than $X$. Their key result states that for any concave utility function $U$, $X \le_r Y \implies \mathbb{E}\left[U(X)\right] \ge \mathbb{E}\left[U(Y)\right]$. The converse implication holds for convex functions. Since the European call payoff is convex in $S$. Hence, when volatility increases through a mean-preserving spread as in the Black-Scholes lognormal model, the expected call payoff necessarily increases with variance. This conclusion is also explicitly stated in standard texts. For example, Shreve~\cite{shreve2004stochastic} notes that as volatility increases, option prices increase in the Black-Scholes-Merton model. 

To make the above convex-order and economic arguments fully explicit within a concrete asset-pricing model, we now establish the monotonic dependence of the European call price on the volatility parameter under the Black-Scholes-Merton framework. This result provides a direct analytical link between variance and call-option valuation and formally justifies the use of variance as a proxy objective in the present setting.
\begin{lemma}
\label{lem:bs_monotonicity}
Under the Black-Scholes model with no dividends and constant risk-free rate $r$, the price of a European call option $C(t,x;\sigma)$ is a strictly increasing function of the volatility parameter $\sigma$. In particular, for fixed $(t,x,K,r,T)$,
\begin{align*}
    \sigma_x > \sigma_y \implies C\left(t,x;\sigma_x\right) > C\left(t,x;\sigma_y\right)
\end{align*}
\end{lemma}
\begin{IEEEproof}
Let $\tau = T - t$ denote the time to maturity. Under the standard Black-Scholes assumptions of no dividends and a constant risk-free interest rate $r$, the price of a European call option at time $t$, with underlying price $S_t = x$, strike $K$, and volatility $\sigma$, is given by
\begin{align*}
    C(t,x;\sigma) = x \Phi(d_1) - K e^{-r\tau} \Phi(d_2),
\end{align*}
where
\begin{align*}
    d_1 = \frac{\ln(x/K) + (r + \tfrac12 \sigma^2)\tau}{\sigma\sqrt{\tau}}, \qquad d_2 = d_1 - \sigma\sqrt{\tau},
\end{align*}
and $\Phi(\cdot)$ and $\phi(\cdot)$ denote the standard normal cumulative distribution function and density, respectively. To examine the sensitivity of the call price with respect to volatility, we compute the derivative $\partial C / \partial \sigma$, commonly referred to as the \textbf{vega}:
\begin{align*}
    \frac{\partial C}{\partial \sigma} = x \phi(d_1) \frac{\partial d_1}{\partial \sigma} - K e^{-r\tau} \phi(d_2) \frac{\partial d_2}{\partial \sigma}.
\end{align*}
Let $A = \ln(x/K) + r\tau$. Then $d_1 = \frac{A}{\sigma\sqrt{\tau}} + \tfrac12 \sigma\sqrt{\tau}$, which implies
\begin{align*}
    \frac{\partial d_1}{\partial \sigma} = -\frac{A}{\sigma^2\sqrt{\tau}} + \tfrac12 \sqrt{\tau}, \qquad \frac{\partial d_2}{\partial \sigma} = -\frac{A}{\sigma^2\sqrt{\tau}} - \tfrac12 \sqrt{\tau}.
\end{align*}
Using the Black-Scholes identity $x \phi(d_1) = K e^{-r\tau} \phi(d_2)$, we obtain
\begin{align*}
    \frac{\partial C}{\partial \sigma} = x \phi(d_1) \left( \frac{\partial d_1}{\partial \sigma} - \frac{\partial d_2}{\partial \sigma} \right) = x \phi(d_1) \sqrt{\tau}.
\end{align*}
Hence
\begin{align*}
    \frac{\partial C}{\partial \sigma} = x \sqrt{\tau} \phi(d_1) > 0.
\end{align*}
Since $x>0$, $\tau>0$, and $\phi(d_1)>0$, the derivative is strictly positive for all admissible parameter values. Therefore, for fixed $(t, x, K, r)$ and $T$, the Black-Scholes call price is a strictly increasing function of the volatility $\sigma$. In particular, if $\sigma_x > \sigma_y$, then $C(t,x;\sigma_x) > C(t,x;\sigma_y)$, which provides a direct analytical confirmation of the general convex-order argument.
\end{IEEEproof}
It is important to clarify that our objective here is not to conflate variance with payoff. Rather, we employ variance as a proxy that captures the characteristics inherent to a call option's utility without treating it as a direct optimization target. This case study serves as a qualitative illustration, demonstrating how variance-aware arm selection can be meaningfully integrated into a real-world decision-making pipeline. Crucially, we do not claim that \texttt{UCB-VV}  or \texttt{SHVV} optimizes option profit; rather, they serve as a variance-driven screening mechanism to identify assets with high potential for favorable price excursions.

\begin{figure}[t]
    \centering
    \includegraphics[width=0.55\linewidth]{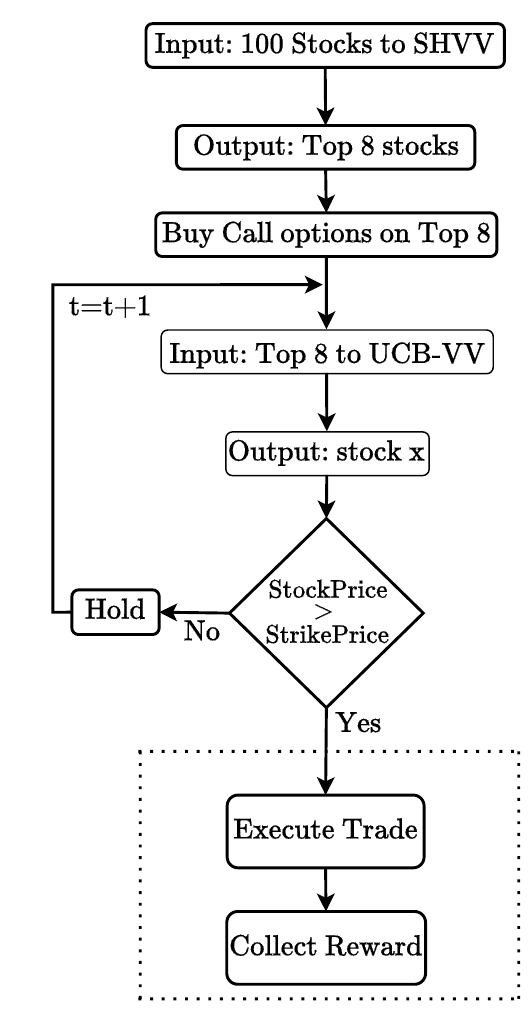} 
    \caption{Flow Chart for call option and trade execution using \texttt{SHVV} and \texttt{UCB-VV}.}
    \label{fig:fig_A}
\end{figure}

Our setting is based on a straightforward premise: investors/buyers of call options favor \say{good uncertainty} as it enhances the potential for significant gains, but are averse to \say{bad uncertainty}, which raises the risk of substantial losses. We define \say{good uncertainty} as volatility linked to positive stock market returns and \say{bad uncertainty} as volatility tied to negative returns. The arm is played only if $S_t > R$ and the buyer profits. The formulation of a call option provides protection against negative returns. We neglect all transaction costs and taxes. Fig. \ref{fig:fig_A} represents the flowchart for using \texttt{UCB-VV} and \texttt{SHVV} for call option trading of \ac{GBM} simulated daily returns for $100$ stocks.

\textbf{Experiment:} We employ \ac{GBM} to simulate stock prices, a widely adopted stochastic process in quantitative finance for its ability to model realistic stock price trajectories, making it invaluable for back-testing trading strategies, evaluating investment opportunities, and pricing financial derivatives. We implement \texttt{SHVV} algorithm to shortlist \enquote{top 8} stocks referring to the $8$ stocks with the highest variances given a budget of $n_1=20e3$. We then employ \texttt{UCB-VV} to select a stock for trading at time $t$ with an additional budget of $n_2=1000$. If the condition $S_t > R$ is met, we collect the corresponding profits. This process continues iteratively: at each time step, we select the next stock according to the \texttt{UCB-VV} index, execute the trade if the condition holds, and accumulate the profits across all selected stocks until time horizon $n=n_1+n_2$.
We compare the performance of our proposed framework with a two stage baseline i.e, \texttt{SH-Mean} followed by \texttt{UCB1} under same budget restrictions. \texttt{SH-Mean} selects the top-8 arms with the highest empirical mean (analogous to \texttt{SHVV}). We see in Fig~2, \texttt{SHVV} + \texttt{UCB-VV} achieve substantially higher cumulative rewards compared to \texttt{SH-Mean} + \texttt{UCB1}.

\begin{figure}[t]
    \centering
    \includegraphics[width=0.65\linewidth]{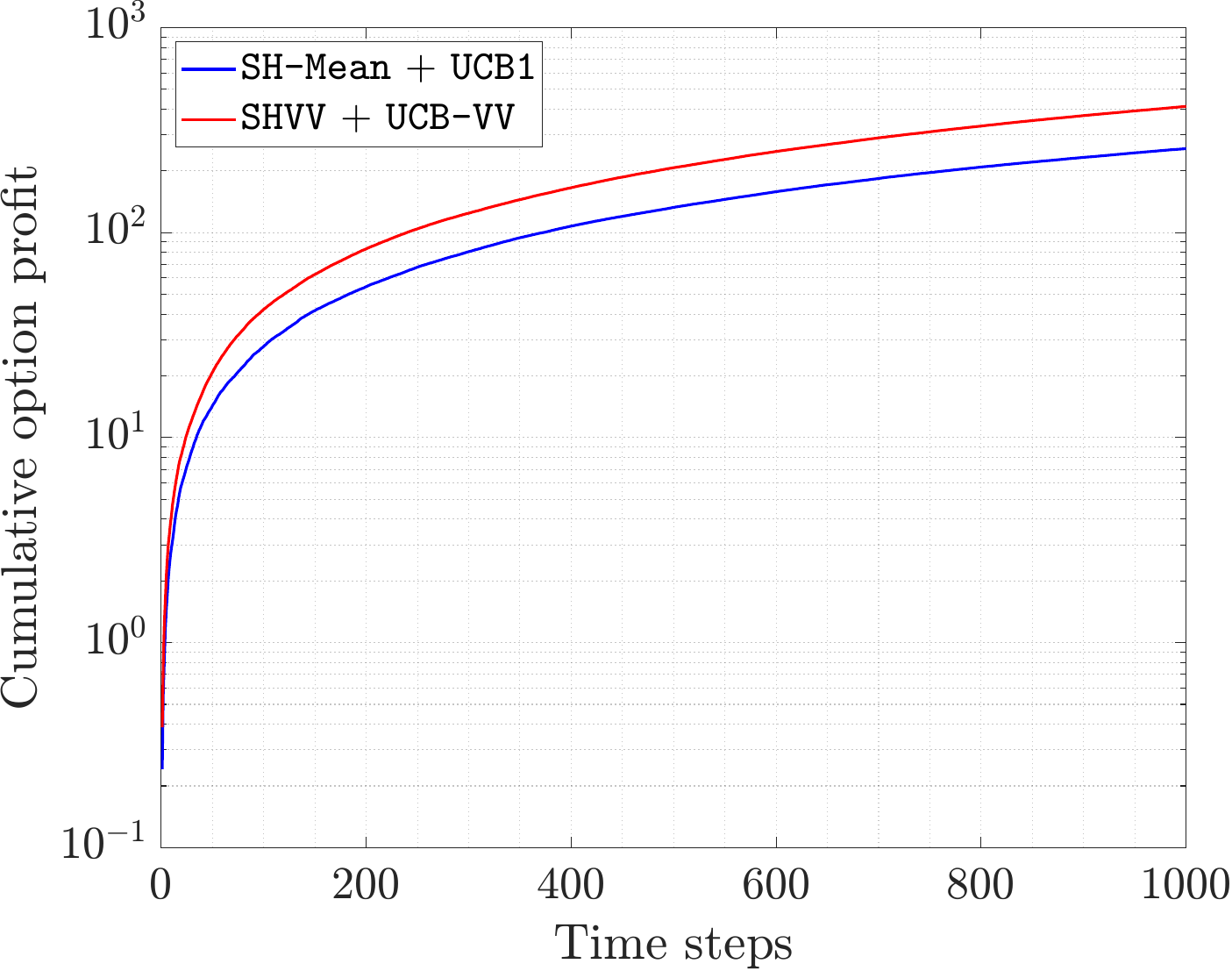}
    \caption{Cumulative profit of \ac{GBM} simulated $100$ stocks for the proposed two stage strategy i.e, \texttt{SHVV} + \texttt{UCB-VV} v/s \texttt{SH-Mean}+ \texttt{UCB1}.}
    \label{fig:fig_exp}
\end{figure}

\section{Future work}
While this work focuses on the stochastic setting with stationary rewards, several promising extensions arise naturally from both theoretical and practical perspectives. First, a minimax formulation of variance bandits where the goal is to minimize misallocation uniformly over a class of distributions would be particularly relevant in adversarial or distribution-free environments where the variance gap is unknown or potentially vanishing. Second, real-world financial time series are inherently non-stationary, with volatility evolving dynamically over time. Extending \texttt{UCB-VV} and \texttt{SHVV} to non-stationary settings via sliding windows, change-point detection, or weighted estimators would significantly enhance their applicability to live trading systems. These directions not only align with the broader vision of robust, risk-aware decision-making but also open new avenues for bridging bandit theory with practical financial engineering.

\section{Appendices}
\subsection{Proof of Theorem 1}
\label{App:1}
We need to bound each $s_i(n)$ to bound the misallocation. Let $ C_{t,s_*} =  \sqrt{\frac{2\log{t}}{s_*}}$ be upper confidence bound. Let $z$ be any arbitrary positive integer and assume that the arm $i$ has been pulled $z$ times, including the pilot fraction. We use a subscript $*$ to any quantity that refers to the optimal arm. At the start of the experiment, each arm is pulled at least once since $S_{\rm P} = K/n$, including the sub-optimal arms. Thus,
\begin{align*}
    s_i(n) &= 1 + \sum_{t= K + 1}^{n} \mathbb{I}\left\{\pi(t) = i\right\} 
\end{align*}
where policy $\pi$ is \texttt{UCB-VV}, and $\mathbb{I}(\cdot)$ is the indicator function:
\begin{align*}
    \mathbb{I} \left\{\pi(t)=i \right\} =
    \begin{cases}
        1 & \mathrm{if}\; \pi(t) = i,\\
        0 & \mathrm{else.}
    \end{cases}
\end{align*}
\begin{align}
 \text{Now}, \;\;   s_i(n) &\overset{(a)} \leq z+ \sum_{t= K + 1}^{n} \mathbb{I}\left\{\pi(t) = i, s_i \geq z \right\} \nonumber \\
    &\overset{(b)} \leq z+ \sum_{t= K + 1}^{n}  \mathbb{I} \Big\{\bar{V}_*(t-1)  + C_{t-1,s_*} \leq \nonumber\\
    &\hspace*{2cm} \bar{V}_i(t-1) + C_{t-1,s_i} ,  s_i \geq z \Big\} \nonumber \\
    &\overset{(c)}\leq z+ \sum_{t= K + 1}^{n}  \mathbb{I} \Big\{\min\limits_{0 < s_* < t} \{\bar{V}_*(s_*) + C_{t-1,s_*} \}  \leq \nonumber\\
    &\hspace*{2cm}\max\limits_{z < s_i < t} \{\bar{V}_{i}(s_i) + C_{t-1,s_i}\}\Big\} \nonumber\\
    &\overset{(d)}\leq z+ \sum_{t= 1}^{\infty}  \sum_{s_*= 1}^{t-1}  \sum_{s_i= z}^{t-1}  \mathbb{I} \Big\{ \bar{V}_*(s_*) + C_{t,s_*} \leq \nonumber\\
    &\hspace*{2cm} \bar{V}_{i}(s_i) + C_{t,s_i}\Big\} 
    \label{eq:six_b}
\end{align}
In step (a), we consider the sub-optimal arm $i$ has been pulled at least $z$ times. In step (b), to pull the sub-optimal arm at time $t$, its index must be at least equal to or greater than that of the optimal arm. Step (c) involves looking back at all the optimal values we had until now, then taking their minimum, and all sub-optimal values we had, then taking their maximum. This will lead to an overcount of (b). In step (d), we upper bound by summing over all indices for the optimal and sub-optimal arms, which will include the event in step (c). Now $\bar{V}_*(s_*) + C_{t,s_*} \leq \bar{V}_{i}(s_i) + C_{t,s_i}$ implies at least one of the following must be true
\begin{align}
    \bar{V}_*(s_*)  \leq \sigma^2_{*} - C_{t,s_*} \label{eq:seven} \\ 
    \bar{V}_{i}(s_i) \geq \sigma_i^2 + C_{t,s_i} \label{eq:eight} \\
    \delta_i = \sigma_*^2 - \sigma_i^2 \leq C_{t-1,s_i} + C_{t-1,s_*}, \label{eq:nine}
\end{align}
$C_{t,s_*}$ and $C_{t,s_i}$ are the bias terms for the optimal and sub-optimal arm, respectively.

We bound the probability of events (\ref{eq:seven}) and (\ref{eq:eight}) using Lemma~\ref{le:le1}:
\begin{align*}
    \mathbb{P} \left(\bar{V}_*(s_*) \leq \sigma^2_{*}-  C_{t,s_*}\right) \leq e^{-2s_*(C_{t,s_*})^2}
\end{align*}
\normalsize
where  $ C_{t,s_*} = \sqrt{\frac{2\log{t}}{s_*}}$. Substitute $C_{t,s_*}$ above, we get
\begin{align*}
    \mathbb{P} \left(\bar{V}_*(r) \leq \sigma^2_{*}-  C_{t,s_*}\right) &\leq e^{-2s_*\left(\sqrt{\frac{2\log{t}}{s_*}}\right)^2} = t^{-4}. 
\end{align*}
Similarly for $C_{t,s_i}$, probability of (\ref{eq:eight}) will be $\leq t^{-4}$. We see that the condition in equation (\ref{eq:nine}) arises when the true variances of the optimal arm and sub-optimal arm are close to each other. Then, the agent is bound to make an error until enough number of pulls have taken place. We split the analysis into two cases: 
\begin{itemize}
    \item \textbf{Case 1:} $s_* \geq z$; in this case, $C_{t,s_*} \leq C_{t,z}$, and we bound $C_{t,s_i} + C_{t,s_*} \leq 2 C_{t,z}$.
    \item \textbf{Case 2:} $s_* < z$; this implies the optimal arm has been pulled fewer than $z$ times. But since the algorithm includes an initial exploration phase (each arm pulled at least once) and $s_*$ grows over time, this case can only occur for a limited number of time steps. We bound the total contribution of this case by a constant.
\end{itemize}
By choosing $z = \left\lceil \frac{8 \log n}{\delta_i^2} \right\rceil$, we ensure that $C_{t,s_i} + C_{t,s_*} < \delta_i$. Thus, for $z = \Big\lceil\frac{8\log{t}}{\delta^2_i}\Big\rceil$, i.e., $s_i = z$, the number of pulls of the sub-optimal arm is substantial enough to make the occurrence of an error event to become highly unlikely. This can be seen as,
\begin{align*}
    \sigma^2_{*} - \sigma_i^2 - 2 C_{t,s_i} &= \sigma^2_{*} - \sigma_i^2 - 2\sqrt{\frac{2\log{t}} {s_i}}\\
    &\hspace*{-2cm}\geq \sigma^2_{*} - \sigma_i^2 -2\sqrt{\frac{2\log{t} \delta^2_i}{8 \log{t}}} = \sigma^2_{*} - \sigma_i^2 - \delta_i = 0.
\end{align*}
It implies that in total of $n$ pulls at a certain time-step $t$, pulls of arm $i$ i.e $s_i \geq \frac{8\log{n}}{\delta^2_i}$ to bound the expected number of pulls for sub-optimal arm $i$ as,
\begin{align*}
    &\mathbb{E}[s_i(n)] \leq \bigg\lceil\frac{8\log{n}}{\delta^2_i}\bigg\rceil + \sum_{t= 1}^{\infty}  \sum_{s_*= 1}^{t-1}  \sum_{s_i= \frac{8\log{n}}{\delta^2_i}}^{t-1} 2 t^{-4} \\ 
    &\leq \frac{8\log{n}}{\delta^2_i} + \sum_{t= 1}^{\infty}  \sum_{s_*= 1}^{t}  \sum_{s_i= 1}^{t} 2 t^{-4} \leq \frac{8\log{n}}{\delta^2_i} + 1 + \frac{\pi^2}{3}.
\end{align*}
Therefore, using union bound
\begin{align*}
     \mathcal{M}_{\texttt{VV}}(n) \leq 8 \sum_{i: \sigma_i^2<\sigma_*^2}\frac{\log{n}}{\delta_i^2} + 1 + \frac{\pi^2}{3}.
\end{align*}
Note that for $s_i \geq {\frac{8\log{n}}{\delta^2_i}}$, the probability of error diminishes at a rate that contributes to \textbf{logarithmic misallocation}.

\subsection{Proof of Lemma 2}
\label{App:2}
Assume a distribution model $\mathcal{F}$, such that the first arm is optimal, denoted by $\ast$. We construct a new distribution model $\mathcal{F}^{i}$ by changing $f_i$ to $\tilde{f}_i$ where $i\neq \ast$ is optimal. Thus 
\begin{align*}
    \mathcal{F}= (f_1, f_2, ....f_i...f_K) \quad \text{and} \quad
    \mathcal{F}^i = (f_1, f_2, ....\tilde{f}_i....f_K)
\end{align*}
such that $\left(\zeta >\tilde{\sigma}_{i}^2- \sigma^2_{*} > 0\right)$ and 
$\left|I(f_i, f_{*})- I(f_i,\tilde{f}_i)\right| \leq \epsilon$ for arbitrary small $\epsilon$ using assumption 1.
Let $\gamma$ be the log-likelihood ratio between the $\mathcal{F}$ and $\mathcal{F}^i \colon \gamma = \log \frac{f_i \left(X_i \left(t_i (1)\right)\right) \dots f_i \left(X_i \left(t_i (s_i) \right)\right)}{\tilde{f}_i \left(X_i \left(t_i (1)\right)\right) \dots \tilde{f}_i \left(X_i \left(t_i (s_i) \right)\right)}$ where $t_i(s_i)$ is the time step when the $s_i$-th observation of arm $i$ occurred. We show that it is unlikely to have $s_i < \frac{C_1 \log n}{I (f_i ,\tilde{f}_i)}$ under two different scenarios for $\gamma$, i.e, $\gamma > C_3 \log n$ and  $\gamma < C_3 \log n$. Note that our proof follows a similar structure to the one presented in \cite{vakili2015mean}, with appropriate modifications to apply it in the context of variance.

\textbf{Case 1}: $\gamma > C_3 \log n$ for a constant $C_3>C_1$.
We have
\begin{align*}
    &\mathbb{P}_\mathcal{F}\bigg[s_i <  \frac{C_1 \log n}{I(f_i,\tilde{f}_i)}, \gamma > C_3 \log n\bigg]\\
    &=  \mathbb{P}_\mathcal{F}\bigg[s_i <  \frac{C_1 \log n}{I(f_i,\tilde{f}_i)}, \sum_{\tau=1}^{s_i} \log \frac{f_i (X_k(\tau))}{\tilde{f}_i(X_k(\tau))} > C_3 \log n\bigg]\\
    &\leq  \mathbb{P}_\mathcal{F} \bigg[ \max_{t\leq \frac{C_1 \log n}{{I(f_i,\tilde{f}_i)}}}\sum_{\tau=1}^{t} \log\frac{f_i (X_k(\tau))}{\tilde{f}_i(X_k(\tau))}> \frac{C_3 \log n}{I(f_i,\tilde{f}_i)} I(f_i,\tilde{f}_i)\bigg] \\ 
    &\leq  \mathbb{P}_\mathcal{F} \Bigg[ \max_{t\leq \frac{C_1 \log n}{{I(f_i,\tilde{f}_i)}}} \frac{1}{t}\sum_{\tau=1}^{t} \log\frac{f_i (X_k(\tau))}{\tilde{f}_i(X_k(\tau))}> \frac{C_3}{C_1}{I(f_i,\tilde{f}_i)} \bigg] 
\end{align*}
By the strong law of large numbers $\frac{1}{t}\sum_{\tau=1}^{t} \log\frac{f_i (X_k(\tau))}{\tilde{f}_i(X_k(\tau))}
\rightarrow  I(f_i,\tilde{f}_i) $ a.s as $t \rightarrow\infty$. Thus, $\max_{t\leq \frac{C_1 \log n}{{I(f_i,\tilde{f}_i)}}} \frac{1}{t}\sum_{\tau=1}^{t} \log\frac{f_i (X_k(\tau))}{\tilde{f}_i(X_k(\tau))} \rightarrow I(f_i,\tilde{f}_i)$ a.s as $n \rightarrow \infty$. So, we have
\begin{align*}
    \mathbb{P}_\mathcal{F} \left[ \max_{t\leq \frac{C_1 \log n}{{I(f_i,\tilde{f}_i)}}} \frac{1}{t}\sum_{\tau=1}^{t} \log\frac{f_i (X_k(\tau))}{\tilde{f}_i(X_k(\tau))}> \frac{C_3}{C_1}{I(f_i,\tilde{f}_i)} \right] \rightarrow 0
\end{align*}
as $n \rightarrow \infty$.

\textbf{Conclusion:} For  $\gamma > C_3 \log n$, by strong law of large numbers, we have:
\begin{align}
    \mathbb{P}_\mathcal{F}\bigg[s_i <  \frac{C_1 \log n}{I(f_i,\tilde{f}_i)}, \gamma > C_3 \log n\bigg] \rightarrow 0,\,  \text{as}\, n \rightarrow \infty
    \label{eq:case1}
\end{align}
Also, using Chernoff bound, $\mathbb{P}_\mathcal{F}[s_i <  \!\!\frac{C_1 \log n}{I(f_i,\tilde{f}_i)}, \gamma > C_3 \log n]$ is upper bounded  for finite $n$ as 
\begin{align}
      \frac{C_1 \log n}{I(f_i,\tilde{f}_i)} n^{-a I(f_i,\tilde{f}_i )
      \frac{(C_3-C_1) ^2}{C_1}}
      \label{eq:upper}
\end{align}
Here $a$ is the Chernoff bound constant. {This result, (\ref{eq:upper}) is obtained from (Eq.(52) of \cite{vakili2015mean})} and is given below.{
\begin{align}
     &\mathbb{P}_\mathcal{F}\bigg[s_i <  \frac{C_1 \log n}{I(f_i,\tilde{f}_i)}, \gamma > C_3 \log n \bigg] \nonumber\\
     &=  \mathbb{P}_\mathcal{F}\bigg[s_i <  \frac{C_1 \log n}{I(f_i,\tilde{f}_i)}, \sum_{s=1}^{s_i} \log \frac{f_i(X_i(s))}{\tilde{f}_i(X_i(s))} > C_3 \log n\bigg]\nonumber\\
     &\leq \mathbb{P}_\mathcal{F} \bigg[\max_{t < \frac{C_1 \log n}{I(f_i,\tilde{f}_i)}} \sum_{s=1}^{t} \log \frac{f_i(X_i(s))}{\tilde{f}_i(X_i(s))} > C_3 \log n\bigg]\nonumber\\
     &\leq \sum_{t=1}^{\frac{C_1 \log n}{I(f_i,\tilde{f}_i)}}\mathbb{P}_\mathcal{F} \bigg[ \frac{1}{t} \sum_{s=1}^{t} \log \frac{f_i(X_i(s))}{\tilde{f}_i(X_i(s))} - \frac{1}{t} C_1 \log {n} >\nonumber\\
     &\hspace*{2.5cm} \frac{1}{t} C_3 \log {n} - \frac{1}{t} C_1 \log {n}\bigg] \nonumber\\
     &\leq \sum_{t=1}^{ \frac{C_1 \log n}{I(f_i,\tilde{f}_i)}}\mathbb{P}_\mathcal{F} \bigg[ \frac{1}{t} \sum_{s=1}^{t} \log \frac{f_i(X_i(s))}{\tilde{f}_i(X_i(s))} - I(f_i,\tilde{f}_i) > \nonumber\\
     &\hspace*{2.5cm} \frac{1}{t} C_3 \log {n} - \frac{1}{t} C_1 \log {n}\bigg]\nonumber\\
     &\leq \sum_{t=1}^{ \frac{C_1 \log n}{I(f_i,\tilde{f}_i)}} \exp \left(\frac{-a_1(C_3-C_1)^2 \log^2 n }{t}\right) \label{eq:A}\\ 
     &\leq \frac{C_1 \log n}{I(f_i,\tilde{f}_i)} n^{-a_1 I(f_i,\tilde{f}_i )
      \frac{(C_3-C_1) ^2}{C_1}}\label{eq:B}
\end{align}
(\ref{eq:A}) holds because  $I(f_i,\tilde{f}_i) >  \frac{1}{t} C_1 \log {n}$ and  (\ref{eq:B}) holds according to Chernoff bound.}

\textbf{Case 2}: $\gamma \leq C_3 \log n$.
By Markov's Inequality, we have 
\begin{align}
    \mathbb{P}_{\mathcal{F}^i}\left[s_i <  \frac{C_1 \log n}{I(f_i,\tilde{f}_i )}\right] &=  \mathbb{P}_{\mathcal{F}^i} \left[n- s_i \geq n - \frac{C_1 \log n}{I(f_i,\tilde{f}_i)}\right] \nonumber\\
    &\leq \frac{\mathbb{E}_{\mathcal{F}^i}[n - s_i]}{n-\frac{C_1 \log n}{I(f_i,\tilde{f}_i)}}
    \label{eq:markov}
\end{align} 
\begin{align}
 \text{Also,} \;\;    \mathbb{P}_{\mathcal{F}}[S(n)] = \mathbb{E}_{\mathcal{F}^i}[\mathbb{I}_{S(n)} e^\gamma],
     \label{eq:measure}
\end{align}
where $S(n)$ is the set of all the observations over $n$ that satisfy a particular event and $\gamma$ is the log-likelihood ratio between $\mathcal{F}$ and $\mathcal{F}^i$. Finally, {(\ref{eq:measure}) is an application of Radon-Nikodym theorem \cite{bain2009fundamentals}.} Using (\ref{eq:markov}) and (\ref{eq:measure}),
\begin{align}
     &\mathbb{P}_\mathcal{F}\left[s_i <  \frac{C_1 \log n}{I(f_i,\tilde{f}_i)}, \gamma \leq C_3 \log n \right] \nonumber\\
     &= \mathbb{E}_\mathcal{F}\left[\mathbb{I}\left\{s_i <  \frac{C_1 \log n}{I\left(f_i,\tilde{f}_i\right)}, \gamma \leq C_3 \log n \right\}\right]\nonumber\\
     &=  \mathbb{E}_{\mathcal{F}^i}\left[\mathbb{I}\left\{s_i <  \frac{C_1 \log n}{I\left(f_i,\tilde{f}_i\right)}, \gamma \leq C_3 \log n \right\} \times e^\gamma \right] \nonumber \\
     &\overset{(a)}{\leq} n ^{C_3} \mathbb{E}_{\mathcal{F}^i} \left[\mathbb{I}\left\{s_i <  \frac{C_1 \log n}{I(f_i,\tilde{f}_i)}, \gamma \leq C_3 \log n \right\} \right] \nonumber\\
     &\leq  n ^{C_3} \mathbb{P}_{\mathcal{F}^i}\left[{s_i < \frac{C_1 \log n}{I\left(f_i,\tilde{f}_i\right)}} \right] \nonumber \leq \frac { n ^{C_3} \mathbb{E}_{\mathcal{F}^i}\left[n-s_i\right]}{n- \frac{C_1 \log n}{I(f_i,\tilde{f}_i)}} \nonumber\\
     &\leq \frac {Kn^{C_3+\alpha}}{{n- \frac{C_1 \log n}{I(f_i,\tilde{f}_i)}}}
     \label{eq:case 2}
\end{align}
Step (a) follows from substituting $\gamma \leq C_3 \log{n}$ in $e^\gamma$. The last inequality is due to $\alpha$-consistency. From \eqref{eq:case1}, \eqref{eq:case 2} and the fact that $ {I(f_i,f_{*})}- {I(f_i,\tilde{f}_i)}$ can be arbitrarily small, for $C_3<1-\alpha$, we have,
\begin{align*}
   \mathbb{P}_\mathcal{F}\bigg[s_i <  \frac{C_1 \log n}{I(f_i,f_{*})}\bigg] \rightarrow 0 \,  \text{as}\, n \rightarrow \infty 
\end{align*}
\begin{align*}
\text{Equivalently,} \;   \mathbb{P}_\mathcal{F}\bigg[s_i \geq \frac{C_1 \log n}{I(f_i,f_{*})}\bigg] \rightarrow 1 \,  \text{as}\, n \rightarrow \infty 
\end{align*}
Using \eqref{eq:upper} and \eqref{eq:case 2}, we conclude, for $C_3 < 1 -\alpha$, when Assumption 2 is satisfied, 
\begin{align*}
    \mathbb{P}_\mathcal{F}\bigg[s_i <  \frac{C_1 \log n}{I(f_i,f_{*})}\bigg] &\leq \frac{C_1 \log n}{I(f_i,\tilde{f}_i)} n^{-a I(f_i,\tilde{f}_i )\frac{(C_3-C_1) ^2}{C_1}} + \\
    &\hspace{2cm} \frac {Kn^{C_3+\alpha}}{{n- \frac{C_1 \log n}{I(f_i,\tilde{f}_i)}}}.
\end{align*}
Thus, there is a  $n_0$ such that for $n>n_0$, $\mathbb{P}_\mathcal{F}\bigg[s_i \geq \frac{C_1 \log n}{I(f_i,f_{*})}\bigg] \geq C_2$, for some constant $C_2 > 0$ independent of $n$ and $\mathcal{F}$. We emphasize that constant $C_1$ and $C_3$ are chosen to satisfy $C_1 < C_3 < 1-\alpha$.

\subsection{Proof of Theorem 2}
\label{App:3}
We need to establish a lower bound on expected sub-optimal pulls, i.e, $ \mathbb{E}[s_i]$, which is a straightforward consequence of Lemma 2. By using Markov's inequality, we have
\begin{align*}
    \mathbb{E}[s_i] \geq \mathbb{P}\left[ s_i \geq \frac{C_1 \log n}{I(f_i,f_{*})}\right]\frac{C_1 \log n}{I(f_i,f_{*})}
\end{align*}
\begin{align*}
\text{So,} \;\;    \lim_{n \rightarrow \infty} \frac{\mathbb{E}[s_i]}{\log n} & \geq  \lim_{n \rightarrow \infty}\mathbb{P}\left[ s_i \geq \frac{C_1 \log n}{I(f_i,f_{*})}\right]\frac{C_1 }{I(f_i,f_{*})} \\
    &=\frac{C_1}{I(f_i,f_{*})}
\end{align*}
Similarly, by Markov's inequality, we have, when assumption 2 is satisfied, there is $n_0 \in n$ such that for all $n > n_0$, 
\begin{align*}
    \mathbb{E}[s_i] \geq \frac{C_1C_2 \log n}{I(f_i,f_{*})} \\
\text{We know,} \;\;    \mathcal{M}(n) = \sum_i^K \mathbb{E}[s_i]  &\geq \sum_{i=1, i\neq *}^K \frac{C_1 C_2\log n}{I(f_i,f_{*})} \\
   \implies \; \liminf_{n \rightarrow \infty} \frac{\mathcal{M}(n)}{\log n} &\geq \sum_{i=1, i\neq *}^K \frac{C_1C_2 }{I(f_i,f_{*})}
\end{align*}
Therefore, the lower bound  is \textbf{order-optimal}.

\subsection{Proof of Lemma 4}
\label{App:4}
$A_r^\prime$ is the set of arms in $A_r$, excluding the $\frac{1}{4} |A_r|= \frac{K}{2^{k+2}}$ arms with the largest variance and letting $D_r$ denote the number of arms in $A_r^\prime$ whose empirical variance is larger than that of the optimal arm,
\begin{align*}
    \mathbb{E}[D_r] &= \sum_{i \in A_r^\prime} \mathbb{P}(\bar{V}_1(t_r)<\bar{V}_i(t_r))
       \leq \sum_{i \in A_r^\prime} \exp\left(-\frac{(t_r-1)^2 \delta_i^2}{2t_r}\right)\\
     &\leq  |A_r^\prime|\max _{i \in A_r^\prime} \exp\left(-\frac{\left(\frac{n} 
         {|A_r|\log_2(K)}-1\right)^2 \delta_i^2}{\frac{2n}{|A_r|\log_2(K)} }\right)\\
     &= |A_r^\prime|\max _{i \in A_r^\prime}   \exp\left(-\frac{({n-|A_r| \log_2(K))^2} \delta_i^2}{2 
         n |A_r| \log_2(K)}\right)\\
     &\leq |A_r^\prime|  \exp\left(-\frac{({n- 4 i_r \log_2(K))^2} \delta_{i_r}^2}{8 n i_r \log_2(K)}\right)
\end{align*}
Using Markov's Inequality, we can bound the probability that $D_r$ exceeds a fraction of $|A_r^\prime|$. Markov's Inequality states that for any non-negative random variable $X$ and constant $a > 0$, $\mathbb{P}(X \geq a) \leq \frac{\mathbb{E}[X]}{a}.$ Therefore,
\begin{align*}
    &\mathbb{P}\left(D_r > \frac{1}{3}|A_r^\prime|\right) \leq 3 \frac{\mathbb{E}[D_r]}{|A_r^\prime|} \leq 3 \mathbb{E}[D_r] \\
    &= 3 \exp \left(\frac{-\left(n-4i_r\log_2(K)\right)^2\delta_{i_r}^2}{ 8 n i_r \log_2(K)}\right).
\end{align*}

\subsection{Proof of Theorem 4}
\label{App:5}
\subsection*{Step 1: Concentration of empirical ${\rm KL}$ divergence}
A bounded distribution is defined on the interval $[l,u]$. The ${\rm KL}$ divergence between 2 such distributions $\nu$ and $\tilde{\nu}$ depends on their PDFs as $I(\nu,\tilde{\nu}) = \int_l^u \log \left(\frac{f_{\nu}(x)}{f_{\tilde{\nu}}(x)} \right) f_{\nu}(x){\rm d}x$. Let $X_i(s)$ denote the samples drawn from arm $i$, which corresponds to arm in the first bandit setting $G^1$. The empirical ${\rm KL}$-divergence is given as:
\vspace*{-0.19cm}
\begin{align*}
    \hat{I}_i(t) = \frac{1}{t}\sum_{s=1}^t \log\left(\frac{f_{\nu_i}(X_i(s))}{f_{\tilde{\nu}_i}(X_i(s))}\right)
\end{align*}
To ensure the empirical ${\rm KL}$ divergence $\hat{I}_i(t)$ concentrates around true ${I}_{i}$, we define a high-probability event $\xi$:
\begin{align*}
    \xi \!\!=\!\! \left\{ \forall 1 \leq i \leq K, 1 \leq t \leq n, \big| \hat{I}_i(t) - {I}_i \big| \!\leq\! \frac{M}{\sqrt{2}}\sqrt{\frac{\log(6nK)}{t}} \right\}.
\end{align*}
This means that for all arms $i$ and all times $t$, the empirical ${\rm KL}$ divergence $\hat{I}_i(t)$  is close to the true ${\rm KL}$ divergence ${I}_{i}$, with a deviation bounded by $ \frac{M}{\sqrt{2}}\sqrt{\frac{\log(6nK)}{t}}$. It ensures that the empirical observations collected by the learner are reliable enough to distinguish between the true distributions. We prove a lemma that $\mathbb{P}_k(\xi)$ is a high-probability event. 

\subsection*{Step 2: It holds that $\mathbb{P}_k(\xi) \geq \frac{5}{6}$.}
To derive the above, we need to upper bound the log-likelihood ratio $\log\left(\frac{f_{\nu_i}(x)}{f_{\tilde{\nu}_i}(x)}\right)$. ${f_{\nu_i}(x)}$ and $f_{\tilde{\nu}_i}(x)$ are PDFs of bounded distributions, we assume, the log-likelihood ratio is also bounded by some constant $M>0$, depending on the specific forms of $f_{\nu_i}(x)$ and  $f_{\tilde{\nu}_i}(x)$. Let $M= \sup_{x\in[l,u]}\left |\log \frac{f_{\nu_i}(x)}{f_{\tilde{\nu}_i}(x)}\right|$. Using lemma \ref{lemma4:BAI_VV}, for $Z_s = \log \frac{f_{\nu_i(x)}}{f_{\tilde{\nu}_i(x)}}, \mathbb{E}[Z_s] = {I}_i$, we can re-write Hoeffding's inequality as $\mathbb{P}\left(\left|\hat{I}_i(t) - {I}_i \geq \epsilon \right| \right) \leq 2\exp\left(\frac{-2{n}\epsilon^2}{M^2}\right)$. We ensure that the KL divergence is also bounded by $M$, from the event $\xi$, defined above.

Let $\epsilon = C_4 \sqrt {\frac{\log (6nK)}{n}}$, where $C_4$ depends on $M$,
\begin{align*}
    &\mathbb{P}\left(\left|\hat{I}_i(t) - {I}_i \right| \geq  C_4 \sqrt {\frac{\log (6nK)}{n}}   \right) \leq \\
    &\hspace*{1cm} 2\exp\left(\frac{-2{t}C_4^2\frac{\log (6nK)}{t}}{M^2}\right)\\
    &\hspace*{1cm}=2\exp\left(\frac{-2C_4^2 \log (6nK)}{M^2}\right)
\end{align*}
Choose $C$ such that $\frac{2C_4^2}{M^2} =1$. Thus, $C_4 = \frac{M}{\sqrt{2}}$. Therefore,
\begin{align*}
    \mathbb{P}\left(\left|\hat{I}_i(t) - {I}_i \right| \geq  \frac{M}{\sqrt{2}} \sqrt {\frac{\log (6nK)}{t}} \right) \leq\frac{1}{6nK} 
\end{align*}
A union bound over all arms and over all time, we get
\begin{align*}
     &\mathbb{P}\left(\exists  i,t : \left|\hat{I}_i(t) - {I}_i \right| \geq  \frac{M}{\sqrt{2}} \sqrt {\frac{\log (6nK)}{t}}  \right)  \leq \frac{Kn}{6nK} = \frac{1}{6}.
\end{align*}
The probability that the event holds is $\mathbb{P}_k({\xi}) \geq 1-\frac{1}{6} = \frac{5}{6}$.

\subsection*{Step 3: Change of Measure}
Let $(n_i)_{1\leq i\leq K}$ denote the numbers of samples collected
by the agent on each arm of the bandits. These quantities are stochastic, but it holds that $\sum_ {1 \leq i \leq K } n_i = n$ by the definition of the fixed budget setting. Let us write for any $0\leq  i \leq K$, $t_i = \mathbb{E}_1 n_i$, where $\mathbb{E}_1 (.)$ denotes the expectation under problem $G^1$. It also holds that $\sum_ {1 \leq i \leq K } t_i = n$. Using the change of measure identity, we relate the probabilities of error under different bandit problems. From \cite{audibert2010best}, we have, for any measurable event $E$ and for any $2 \leq k \leq K$:
\begin{align}
\mathbb{P}_{k}(E) = \mathbb{E}_1\left[\mathbb{I}\{E\} \exp\left(-n_{k} \cdot \hat{{I}}_k(n_k)\right)\right],
\label{eq:Change}
\end{align}
Consider an event $E_{k} = \{J_n= 1\} \cap \{\xi\} \cap \{n_{k} \leq 6t_{k}\}$, where:
\begin{itemize}
    \item $J_n= 1$: The learner outputs arm $1$ as the best arm,
    \item $\xi$: A high-probability event where empirical ${\rm KL}$ divergences concentrate,
    \item $n_{k} \leq 6t_{k}$: Arm $k$ is pulled at most $6t_{k}$ times.
\end{itemize}
From (\ref{eq:Change}), we have
\begin{align}
    \mathbb{P}_{k}(E_k) = \mathbb{E}_1\left[\mathbb{I}\{E\} \exp\left(-n_{k} \cdot \hat{{I}}_k(n_k)\right)\right] \nonumber \\
    \geq  \exp\left[-6t_{k} {I}_k - \frac{M}{\sqrt{2}} \sqrt{n \log( 6nK)}\right] \mathbb{P}_1 [E_k]
    \label{eq:error}
\end{align}

\subsection*{Step 4: Lower Bound on $\mathbb{P}_1 [E_{k}]$.}
Assume that for the algorithm we consider
\begin{align}
    \mathbb{E}_{1}(J_n \neq 1) \leq \frac{1}{2}
    \label{eq:prob}
\end{align}
i.e., that the probability that the algorithm makes a mistake on problem $1$ is less than $\frac{1}{2}$.
For any $2\leq i \leq K$, it holds by Markov's inequality that 
\begin{align}
    \mathbb{P}_1(n_i \geq 6t_i) \leq \frac{\mathbb{E}_1 n_i}{6 t_i} = \frac{1}{6}
    \label{eq:markov1}
\end{align}
Using \eqref{eq:prob} and \eqref{eq:markov1}, Lemma 5 ({$\mathbb{P}_k(\xi) \geq \frac{5}{6}$) 
and using union bounds for any $2 \leq k \leq K$, 
\begin{align*}
    \mathbb{P}_1(E_{k}) \geq 1- \bigg(\frac{1}{6}+ \frac{1}{2}+ \frac{1}{6} \bigg) = \frac{1}{6}.
\end{align*}
Using (\ref{eq:error}) and the fact that for any $2\leq k\leq K, \mathbb{P}_k (J_n \neq k) \geq \mathbb{P}_{k}(E_{k})$ implies that for any $2 \leq k \leq K$
\begin{align*}
   \mathbb{P}_{k} \left({J_n \neq {k}}\right) \geq  \frac{1}{6} \exp\left[-6t_{k} {I}_{k} - \frac{M}{\sqrt{2}} \sqrt{n \log( 6nK)}\right] 
\end{align*}
KL-divergence in case of variance for uniform distribution is:
\begin{align}
  I_k=  \frac{1}{2} \log \frac{f_{\nu}(x)}{f_{\tilde{\nu}}(x)} = \frac{1}{2}\log {\frac{\sigma_{\tilde{\nu}}^2}{\sigma_{\nu}^2}}
    \leq \frac{1}{2} \left(\frac{\sigma_{\tilde{\nu}}^2}{\sigma_{\nu}^2} -1\right) = \frac{\delta_k}{2 \sigma_{\nu}^2}
     \label{eq:KLUni}
\end{align}

\subsection*{Step 5: Final bound}
Since $\sum_{2 \leq i\leq K} \delta_i^{-2} = H (1)$ and $\sum_{1 \leq i \leq K} n_i = n$, then there exists  $2 \leq k \leq K$ such that $ t_k =\frac{n}{ H (1) \delta_k^2}$.
Combining the results from Steps 1--3, we derive the lower bounds on the error probability in terms of $H(1)$, which is given as;
\begin{align*}
   \mathbb{P}_{k} \left({J_n \neq {k}}\right) &\geq  \frac{1}{6} \exp\left[-6  \frac{n}{H(1) \delta_k^2} \frac{\delta_k}{2\sigma_{\nu}^2} - \frac{M}{\sqrt{2}} \sqrt{n \log(6nK)}\right]\\
   &  = \frac{1}{6} \exp\left[-3  \frac{n}{H_\sigma(1) \sigma_{\nu}^2} - \frac{M}{\sqrt{2}} \sqrt{n \log( 6nK)}\right].
\end{align*}
 
\textbf{Note:} For Bernoulli distribution, lower bound proves order-optimality and can be followed from \cite{carpentier2016tight} with some minor changes in bounded support of $\nu_k$ and $\tilde{\nu}_k$.

\subsection{Proof of Theorem 5}
\begin{IEEEproof}
\label{App:6}
Rewriting $ \bar{V}(n) $ in terms of the true mean $\mu$:
\begin{align*}
\bar{V}(n) = {\frac{1}{n} \sum_{i=1}^n (X_i - \mu)^2} - {(\bar{\mu}(n) - \mu)^2}.
\end{align*}
Applying the triangle inequality, we have
\begin{align*}
|\bar{V}(n) - \sigma^2| \leq \left|\frac{1}{n} \sum_{i=1}^n (X_i - \mu)^2 - \sigma^2 \right| + (\bar{\mu}(n) - \mu)^2.
\end{align*}
Let $Y_i = (X_i - \mu)^2$. Since $X_i$ is sub-Gaussian with parameter $v^2$, $Y_i$ is sub-exponential with parameters $ (v^2, 4v^2) $. Applying Bernstein’s inequality for sub-exponential random variables,
\begin{align*}
\mathbb{P}\left(\left|\frac{1}{n} \sum_{i=1}^n Y_i - \sigma^2\right| > \epsilon\right) \leq 2 \exp\left(-C_5 n \min\left(\frac{\epsilon^2}{v^4}, \frac{\epsilon}{v^2}\right)\right),
\end{align*}
for some constant $ C_5 > 0 $. For the second term, since $ \bar{\mu}(n) $ is sub-Gaussian with parameter $ v^2/n $. Therefore,
\begin{align*}
\mathbb{P}\left(|\bar{\mu}(n) - \mu| > \epsilon_1\right) \leq 2 \exp\left(-\frac{n\epsilon_1^2}{2v^2}\right).
\end{align*}
Setting $ \epsilon_1 = \sqrt{\epsilon/2} $, we get
$\mathbb{P}\left((\bar{\mu}(n) - \mu)^2 > \frac{\epsilon}{2}\right) \leq 2 \exp\left(-\frac{n\epsilon}{4v^2}\right).$
Combining both terms:
\begin{align*}
\mathbb{P}\left(|\bar{V}(n) - \sigma^2| > \epsilon\right) \leq 2 \exp\left(-C_5 n \min\left(\frac{\epsilon^2}{v^4}, \frac{\epsilon}{v^2}\right)\right)+
\end{align*}
\begin{align}
&\hspace{3cm}2 \exp\left(-\frac{n\epsilon}{4v^2}\right)
\label{the_5}
\end{align}

Thus, $\exists C_V:0 < C_V < \min\left\{\frac{1}{4}, C_5\right\}$ such that the statement of the theorem follows.
\end{IEEEproof}

\subsection{Proof of Theorem 6}
\begin{IEEEproof}
\label{App:7}
We begin by expressing the deviation of the empirical \ac{SR} from its true value:
\begin{align*}
   |\bar{S}(n) - S| = \left|\frac{\bar{\mu}(n)}{\sqrt{\bar{V}(n)}} - \frac{\mu}{\sigma}\right| = \left|\frac{\bar{\mu}(n) \sigma - \mu \sqrt{\bar{V}(n)}}{\sqrt{\bar{V}(n)} \sigma}\right|.
\end{align*}
By the triangle inequality, we decompose the equation as,
\begin{align*}
    |\bar{S}(n) - S| \leq \frac{|\bar{\mu}(n) - \mu|}{\sqrt{\bar{V}(n)}} + \frac{|\mu| \cdot |\sqrt{\bar{V}(n)} - \sigma|}{\sqrt{\bar{V}(n)} \sigma}.
\end{align*}
We first leverage concentration inequalities to stabilize the denominator by enforcing its constancy, i.e., to show with high probability that $\bar{V}(n) \geq \frac{\sigma^2}{4}$. Utilizing the result of Corollary 1 we get:
\begin{align*}
    \mathbb{P}\left(\sqrt{\bar{V}(n)} \geq \frac{\sigma}{2}\right) \leq 2 \exp\left(-C n \min\left(\frac{\sigma^2}{2v^4}, \frac{\sigma}{2v^2}\right)\right).
\end{align*}
Thus with probability at least $1 - 2 \exp\left(-C n \min\left(\frac{\sigma^2}{2v^4}, \frac{\sigma}{2v^2}\right)\right) $, we have $\sqrt{\bar{V}(n)} \geq \frac{\sigma}{2} \,\Rightarrow\, \frac{1}{\sqrt{\bar{V}(n)}} \leq \frac{2}{\sigma}$. For this event, the deviation bound becomes: 
\begin{align*}
  |\bar{S}(n) - S| \leq \frac{2|\bar{\mu}(n) - \mu|}{\sigma} + \frac{2|\mu| \cdot |\sqrt{\bar{V}(n)} - \sigma|}{\sigma^2}.
\end{align*}
To ensure $ |\bar{S}(n) - S| \leq \eta $, it suffices to require both terms on the right-hand side to be at most $ \eta/2 $. That is,  
\begin{align*}
    |\bar{\mu}(n) - \mu| \leq \frac{\eta \sigma}{4}, \quad |\sqrt{\bar{V}(n)} - \sigma| \leq \frac{\eta \sigma^2}{4|\mu|}.
\end{align*}  
Using the concentration bound for the sample mean $\mathbb{P}\left(|\bar{\mu}(n) - \mu| \geq \epsilon\right) \leq 2 \exp\left(-\frac{n\epsilon^2}{2v^2}\right)$, we set $ \epsilon = \frac{\eta \sigma}{4} $ to obtain:  
\begin{align*}
    \mathbb{P}\left(|\bar{\mu}(n) - \mu| \geq \frac{\eta \sigma}{4}\right) \leq 2 \exp\left(-\frac{n \eta^2 \sigma^2}{32 v^2}\right).
\end{align*}
Similarly, invoking the sub-exponential tail bound stated in Corollary 1 and setting $\epsilon_1 = \frac{\eta \sigma^2}{4|\mu|}$, we obtain
\begin{align*}
    &\mathbb{P}\left(|\sqrt{\bar{V}(n)} - \sigma| \geq \frac{\eta \sigma^2}{4|\mu|}\right) \leq \\
    &\hspace*{2cm} 2 \exp\left(-C_7 n \min\left(\frac{\eta^2 \sigma^4}{16 |\mu|^2 v^4}, \frac{\eta \sigma^2}{4 |\mu| v^2}\right)\right),
\end{align*}   
for some constant $C_7 > 0$. Combining these bounds via the union bound, the total probability of deviation satisfies:  
\begin{align*}
    \mathbb{P}\left(|\bar{S}(n) - S| \geq \eta\right) &\leq 2 \exp\left(-\frac{n \eta^2 \sigma^2}{32 v^2}\right) + \\
    & 2 \exp\left(-C_7 n \min\left(\frac{\eta^2 \sigma^4}{16 |\mu|^2 v^4}, \frac{\eta \sigma^2}{4|\mu| v^2}\right)\right).
\end{align*}
Each exponential term decays at a rate governed by $n \min(\eta^2, \eta)$, up to constants depending on $\mu, \sigma^2, v^2$. Absorbing all constants into a single parameter $c > 0$, we conclude the statement of the theorem.
\end{IEEEproof}}

\subsection{Proof of Theorem 7}
\begin{IEEEproof}
\label{App:8}
The proof largely follows the corresponding concentration inequality and is in line with other proofs of UCB-type algorithms. Hence, here we provide an abridged sketch of the proof. From the concentration inequality on sample \ac{SR}, the estimated \ac{SR} $\bar{S}_i(t) $ satisfies:
\begin{align*}
   \mathbb{P}\left(|\bar{S}_i(t) - S_i| > \beta_i(t)\right) \leq \frac{1}{t^2},
\end{align*}

Applying a union bound over all arms $i$ up to  $n$:
\begin{align*}
   \mathbb{P}\left(\exists i, t \leq n : |\bar{S}_i(t) - S_i| > \beta_i(t)\right) \leq \sum_{i=1}^K \sum_{t=1}^n \frac{1}{t^2} \leq \frac{\pi^2}{6}K.
\end{align*}
Thus, with high probability, for all $i$ and $t$, we have:
\begin{align*}
   S_i \leq \bar{S}_i(t) + \beta_i(t), \quad \bar{S}_i(t) \leq S_i + \beta_i(t).
\end{align*}

Suppose a suboptimal arm $ i $ (with $ \Delta_i > 0 $) is selected at time $ t $. By the \texttt{UCB-Sharpe} algorithm, its index must satisfy:
\begin{align*}
   \bar{S}_i(t) + \beta_i(t) \geq \bar{S}_{i_*}(t) + \beta_{i_*}(t) \\ \Rightarrow S_i + 2\beta_i(t) \geq S_* - \beta_{i_*}(t)\;   \Rightarrow  \Delta_i \leq 3\beta_i(t)
\end{align*}

Solve for $s_i(t)$ using $\beta_i(t) = \sqrt{\frac{1}{C s_i(t)} \log(4t^2)}$, we get
  \begin{align*}
    \Delta_i \leq 3\sqrt{\frac{1}{C s_i(t)} \log(4n^2)} \quad \Rightarrow \quad s_i(t) \leq \frac{9}{C \Delta_i^2} \log(4n^2).
  \end{align*}
Summing over all suboptimal arms:
   \begin{align*}
      \mathbb{E}[s_i(n)] \leq \frac{9}{C \Delta_i^2} \log(4n^2) + \mathcal{O}(1).
  \end{align*}
Therefore, using (\ref{eq:reg-SR}), we get the statement of the theorem.
\end{IEEEproof}

\bibliographystyle{ieeetr}
\bibliography{references}
\end{document}